\documentclass{article} 
\usepackage[top=3.5cm,right=3cm,bottom=3.5cm,left=3cm]{geometry}
\usepackage[english]{babel} 
\usepackage{amssymb}
\usepackage{amsmath}
\usepackage{txfonts}
\usepackage{mathdots}
\usepackage[classicReIm]{kpfonts}
\usepackage{graphicx}

\begin{document}

\noindent 

\noindent 

\noindent 

\noindent 
\begin{center}
\noindent \textbf{\large A two - phase - ACO algorithm for solving nonlinear optimization problems subjected to fuzzy relational equations}
\end{center}
\bigskip
\noindent \textbf{}
\begin{center}

\noindent \textbf{Amin Ghodousian${}^{a,}$}${}^{}$\footnote{\ Corresponding\ author$  $Email\ addresses:\ \ $  $a.ghodousian@ut.ac.ir\ (Amin\ Ghodousian).}\textbf{, Sara Zal${}^{b}$}

\end{center}
\bigskip
\noindent {\footnotesize ${}^{a,}$Faculty of Engineering Science, College of Engineering, University of Tehran, P.O.Box 11365-4563, Tehran, Iran.\noindent\medskip }

\noindent {\footnotesize ${}^{b}$Department of Engineering Science, College of Engineering, University of Tehran, Tehran, Iran.\textbf{}}

\noindent \textbf{}

\noindent \textbf{Abstract}
\vskip 0.2in
\noindent {\small In this paper, we investigate nonlinear optimization problems whose constraints are defined as fuzzy relational equations (FRE) with max-min composition. Since the feasible solution set of the FRE is often a non-convex set and the resolution of the FREs is an NP-hard problem, conventional nonlinear approaches may involve high computational complexity. Based on the theoretical aspects of the problem, an algorithm (called FRE-ACO algorithm) is presented which benefits from the structural properties of the FREs, the ability of discrete ant colony optimization algorithm (denoted by ACO) to tackle combinatorial problems, and that of continuous ant colony optimization algorithm (denoted by ACO${}_{R}$) to solve continuous optimization problems. In the current method, the fundamental ideas underlying ACO and ACO${}_{R}$ are combined and form an efficient approach to solve the nonlinear optimization problems constrained with such non-convex regions. Moreover, FRE-ACO algorithm preserves the feasibility of new generated solutions without having to initially find the minimal solutions of the feasible region or check the feasibility after generating the new solutions. FRE-ACO algorithm has been compared with some related works proposed for solving nonlinear optimization problems with respect to max-min FREs. The obtained results demonstrate that the proposed algorithm has a higher convergence rate and requires a less number of function evaluations compared to other considered algorithms.}


\noindent 
\vskip 0.2in
\noindent \textbf{1. Introduction}
\vskip 0.2in
\noindent In this paper, we study the following nonlinear optimization problem in which the constraints are formed as the fuzzy relational equations defined by the minimum t-norm:
\begin{equation} \label{(1)} 
\begin{array}{l} {\min \, \, \, \, f({\it x}){\it \; \; }} \\ {\, \, \, \, \, \, \, \, \, \, \, {\it A}\, \varphi \, {\it x}={\it b}} \\ {\, \, \, \, \, \, \, \, \, \, {\it x}\in [0,1]^{n} } \end{array} 
\end{equation} 
where $I=\{ \, 1\, ,\, 2\, ,\, ...\, ,\, m\, \} $ and $J=\{ \, 1\, ,\, 2\, ,\, ...\, ,\, n\, \} $. ${\it A}=(a_{ij} )_{m\, \times n} $ is a fuzzy matrix such that $0\le a_{ij} \le 1$ ($\forall i\in I$ and $\forall j\in J$), ${\it b}=(b_{i} )_{m\times 1} $is a fuzzy vector such that $0\le b_{i} \le 1$ ($\forall i\in I$) and ``$\varphi $'' is the max-min composition, i.e., $\varphi (x,y)=\min \{ x,y\} $. So, if \textbf{${\it a}_{{\it i}} $} ($i\in I$) denotes the $i$`th row of matrix ${\it A}$, then the $i$`th constraint of problem(1) can be expressed as \textbf{${\it a}_{{\it i}} \varphi {\it x}=b_{i} $} where \textbf{${\it a}_{{\it i}} \varphi {\it x}=\max _{j=1}^{n} \left\{\varphi (a_{ij} ,x_{j} )\right\}=\max _{j=1}^{n} \left\{\min \{ \, a_{ij} ,x_{j} \, \} \right\}$}. 

\noindent  The theory of fuzzy relational equations (FRE) with max-min composition was firstly proposed by Sanchez as a generalized version of Boolean relation equations [50]. Besides, he developed the application of max-min FRE in medical diagnosis [52]. Nowadays, it is well-known that many issues associated to the body can be treated as FRE problems [40]. On the other hand, the fundamental result for FRE with max-product composition goes back to Pedrycz [41] and was further studied in [2,32]. Since then, many theoretical improvements have been investigated [20,23,42,47,58] and various types of FRE with continuous triangular norms were used in applied topics [6,39,47]. In [45] the author demonstrated how problems of interpolation and approximation of fuzzy functions are converted with solvability of systems of FRE. Generally, when rules of inference are applied and their corresponding consequences are known, the problem of determining antecedents is simplified and mathematically reduced to solving an FRE [38]. 

\noindent The solvability identification and finding set of solutions are the primary, and the most fundamental, matter concerning the FRE problems. Di Nola et al. proved that the solution set for FRE (if it is nonempty), defined by max-min composition is often a non-convex set. This non-convex set is completely determined by one maximum solution and a finite number of minimal solutions [7]. In fact, the same results hold true for a more general t-norms instead of the minimum operator [17]. On the other hand, Chen and Wang [4] presented an algorithm for obtaining the logical representation of all minimal solutions and deduced that a polynomial-time algorithm with the ability to find all minimal solutions of FRE (with max-min composition) may not exist. Also, Markovskii showed that solving max-product FRE is closely related to the covering problem which is a type of NP-hard problem [36]. Additionally, Lin et al. [29] demonstrated that all systems of max-continuous t-norm fuzzy relational equations, for example, max-product, max-continuous Archimedean t-norm and max-arithmetic mean are essentially equivalent, because they are all equivalent to the set covering problem. Over the past decades, the solvability of FREs which are defined using different max-t compositions have been investigated by many researchers [15,18,43,46,49,54,59,62].  

\noindent The optimization problem related to FRE and FRI is one of the most interesting and on-going research topics among similar problems [3,11,12,16-22,25,31,52,\\*57,60,64]. For example, Fang and Li converted a linear optimization problem subjected to FRE constraints with max-min operation into an integer programming problem and solved it by branch and bound method using jump-tracking technique [12]. Wu et al. worked on improvement of the method employed by Fang and Li; this was done by decreasing the search domain and presented a simplification process by three rules which resulted from a necessary condition [61]. Chang and Shieh [3] presented new theoretical results concerning the same problem by improving an upper bound on the optimal objective value. In [24], an application of optimizing the linear objective with respect to max-min FRE constraints was employed for a minimum cost streaming media provider problem. Linear optimization problem was further investigated by numerous scholars focusing on max-product operation [21,32]. Loetamonphong and Fang defined two sub-problems by separating negative from non-negative coefficients in the objective function, and then obtained an optimal solution by combining the two sub-problems [32]. The maximum solution of FRE is the optimum for the sub-problem having negative coefficients. Another sub-problem was converted into a binary programming problem and was solved using branch and bound method. Moreover, generalizations of the linear optimization problem with respect to FRE have been studied; this was done through replacement of max-min and max-product compositions with different fuzzy compositions [25,52,57].   

\noindent Recently, many interesting generalizations of the linear programming problems constrained by FRE or FRI have been introduced and developed based on composite operations and fuzzy relations used in the definition of the constraints, and some developments on the objective function of the problems [5,9,13,16,26,31,61]. For instance, the linear optimization of bipolar FRE was studied by some researchers where FRE was defined with max-min composition [13] and max-Lukasiewicz composition [26,31]. In [26] the authors introduced the optimization problem subjected to a system of FRE defined as \textbf{$X({\it A}^{+} ,{\it A}^{-} ,{\it b})=\{ {\it x}\in [0,1]^{m} \, :\, {\it x}\circ {\it A}^{+} \vee \tilde{{\it x}}\circ {\it A}^{-} ={\it b}\} $} where \textbf{$\tilde{{\it x}}_{{\it i}} =1-{\it x}_{{\it i}} $ }for each component of \textbf{$\tilde{{\it x}}=(\tilde{x}_{i} )_{1\times m} $ }and the notations ``\textbf{$\vee $}'' and ``\textbf{$\circ $}'' denote max operation and the max-Lukasiewicz composition, respectively. They translated the problem into a 0-1 integer linear programming problem which is then solved using well-developed techniques. In [31], the foregoing problem was solved by an analytical method based on the resolution of the feasible region. In [17], the authors introduced a linear optimization problem with constraints formed by \textbf{$X({\it A},{\it D},{\it b}^{1} ,{\it b}^{2} )=\{ {\it x}\in [0,1]^{n} \, :\, {\it A}\varphi {\it x}\le {\it b}^{1} ,{\it D}\varphi {\it x}\ge {\it b}^{2} \} $} where \textbf{$\varphi $} represents an operator with (closed) convex solutions. This problem was further investigated with $\varphi $as the Dubois-Prade t-norm [14] and $\varphi $as a continuous increasing operator [37]. Yang et al. [66] studied the optimal solution of minimizing a linear objective function with addition-min FRI constraints defined as \textbf{$\sum _{j=1}^{n}\min \{ \, a_{ij} ,x_{j} \, \}  \ge b_{i} $}  for \textbf{$i=1,...,m$}. Ghodousian et al. [16,19] introduced linear and nonlinear optimization problems with FRI fuzzy constraints (FRI-FC). The feasible region of the FRI-FC problems are defined as \textbf{$X({\it A},{\it b})=\{ {\it x}\in [0,1]^{n} \, :\, {\it A}\varphi {\it x}{\rm \preceq b}\} $}, where $\varphi $ is an arbitrary continuous t-norm and `` \textbf{${\rm \preceq }$}''\textbf{ }denotes the relaxed or fuzzy version of the ordinary inequality ``\textbf{$\le $}''. 

\noindent The optimization problems with general nonlinear objective functions and FRE or FRI constraints were studied in [5,9,15,18,27,55,61,64,66]. For example, Wu et al. presented an efficient method to optimize a linear fractional programming problem under FRE with max-Archimedean t-norm composition [62]. Dempe and Ruziyeva [5] generalized the fuzzy linear optimization problem by considering fuzzy coefficients. Dubey et al. studied linear programming problems involving interval uncertainty modeled using intuitionistic fuzzy set [9]. Yang et al. [67] studied the single-variable term semi-latticized geometric programming subject to max-product fuzzy relation equations. The proposed problem was devised from the peer-to-peer network system and the target was to minimize the biggest dissatisfaction degrees of the terminals in such system. Latticied optimization problem was introduced in [56] where the constraints and the objective function were defined as max-min FRI and \textbf{$z({\it x})=\max _{i=1}^{n} \{ \, \min \{ \, c_{i} ,x_{i} \, \} \, \} $ }with \textbf{$c_{i} \in [0,1]$},  respectively. The latticized programming problem subjected to max-min FRI was also investigated in the works [27,64]. In [27], based on the concept of semi-tensor product, a matrix approach was applied to handle the latticized linear programming problem subjected to max-min FRI. Also, Yang et al. [65] introduced another version of the latticized programming problem defined by minimizing  objective function \textbf{$z({\it x})=\max _{j=1}^{n} \{ \, x_{j} \, \} $ }subject to \textbf{$X({\it A},{\it b})=\{ {\it x}\in [0,1]^{n} \, \, :\, \, {\it A}\circ {\it x}\ge {\it b}\} $} where `'\textbf{$\circ $}'' denotes fuzzy max-product composition.\textbf{ }On the other hand, Lu and Fang considered the single non-linear objective function and solved it with FRE constraints and max-min operator [34]. They proposed a genetic algorithm for solving the problem. In [15], the authors improved Lu and Fang's algorithm by introducing different crossover and mutation operators. 

\noindent Generally, there are three important difficulties related to FRE or FRI problems. Firstly, in order to completely determine FREs and FRIs, we must initially find all the minimal solutions, which is an NP-hard problem. Secondly, a feasible region formed as FRE or FRI is often a non-convex set. Thirdly, FREs and FRIs as feasible regions lead to optimization problems with highly non-linear constraints. Due to the above mentioned difficulties, although the analytical methods are efficient to find exact optimal solutions, they may also involve high computational complexity for high-dimensional problems (especially, if the simplification processes cannot considerably reduce the problem). Furthermore, in spite of the fact that deterministic approaches such as feasible direction and generalized gradient descent methods are fast and able to quickly find a local minimum, they make strong assumptions on the continuity and differentiability of the constraints as well as the objective function [1]. On the other hand, heuristic algorithms, like Ant Colony Optimization (ACO) [8,53], do not make such assumptions and they have been successfully applied for tackling constrained optimization problems [15,18,19,28,34,48]. Particularly, we refer the readers to [10,23,30,35,44,63] in which the powerful performance of ant colony optimization algorithm has been shown in applied optimization problems.

\noindent In this paper, an ACO-based algorithm (FRE-ACO) is proposed for solving the nonlinear optimization problems constrained with max-min fuzzy relational equations. The FRE-ACO algorithm consists of two general phases that are sequentially employed until a termination criterion is satisfied. Phase (I) is based on (discrete) ACO and plays a main role~in~exploring~the new areas of the search space. At each iteration, the purpose of Phase (I) is to find some minimal candidate solution(s) for the max-min FRE, that is a combinatorial problem. This is done by selecting FRE-paths that are attained through use of the structural properties of the FREs, and according to their corresponding pheromone values. Given a minimal candidate solution (generated by a FRE-path), the algorithm can find a subset of the feasible region that is analytically a closed convex cell. In fact, the algorithm associates~to~each FRE-path~exactly one~closed convex cell. The second phase minimizes the original objective function starting with the feasible solutions that are randomly selected from the feasible cells obtained at the end of Phase (I). Phase (II) is based on ACO${}_{R}$ and guarantees the exploitation of the algorithm. At the end of Phase (II), the solution archive is used to update the pheromone values of the previously found FRE-paths. Better solutions increase the probability of choosing their corresponding cells at the next iteration, by~increasing the pheromone~on those FRE-paths associated with their cells. So, at the next iteration, Phase (I) is biased towards FRE-paths that are associated with the most promising cells (i.e., those cells including better solutions). In conclusion, Phase (I) provides feasible regions for phase (II) to exploit, and Phase (II) in turn updates the pheromone values of the FRE-paths used by phase (I) for finding better feasible regions. As it will be shown, the proposed algorithm keeps the search inside of the feasible region without checking the feasibility of new generated solutions. 

\noindent The current paper~consists of three main parts. Firstly, we review some structural details of FREs defined by the minimum t-norm such as the theoretical properties of the feasible solution set and some necessary and sufficient conditions for the feasibility of the problem. Then, two preceding phases are described in detail and followed by explanation of FRE-ACO. Finally, some statistical and experimental results are provided to evaluate the performance of our algorithm. Moreover, a comparison is made between the FRE-ACO algorithm and the genetic algorithms presented in [15] and [34] for solving nonlinear optimization problems with respect to max-min FREs. The remainder of the paper is organized as follows. Section 2 takes a brief look at some basic results on the feasible solutions set of problem(1). These results are used throughout the paper and provide a proper background to design an efficient algorithm for solving the problem. The details of Phase (I) and Phase (II) are studied in Sections 3 and 4, respectively. In section 5, the FRE-ACO algorithm and its characteristics are described and a comparative study is presented. Finally, in section 7, the experimental results are demonstrated. 

\noindent 
\vskip 0.2in
\noindent \textbf{2. A review of the structure of max-min FRE}
\vskip 0.1in
\noindent This section briefly describes some relevant results about the solution to system of max-min fuzzy relational equations. Let \textbf{$S({\it A},{\it b})$ }denotes the feasible solution set of problem(1), that is, \textbf{$S({\it A},{\it b})=\{ \, {\it x}\in [0,1]^{n} \, :\, \, {\it A}\, \varphi \, {\it x}={\it b}\, \} $}. For \textbf{${\it x},{\it y}\in [0,1]^{n} $}, we say \textbf{${\it x}\le {\it y}$} if and only if \textbf{${\it x}_{j} \le {\it y}_{j} $}, $\forall j\in J$. Moreover, a solution \textbf{$\bar{{\it x}}\in S({\it A},{\it b})$} is said to be a maximum solution, if \textbf{${\it x}\le \bar{{\it x}}$}, \textbf{$\forall {\it x}\in S({\it A},{\it b})$}. Similarly, \textbf{$\underline{{\it x}}\in S({\it A},{\it b})$} is called a minimal solution, if \textbf{${\it x}\le \underline{{\it x}}$}implies \textbf{${\it x}=\underline{{\it x}}$}, \textbf{$\forall {\it x}\in S({\it A},{\it b})$}. Also, the notation \textbf{$\underline{S}({\it A},{\it b})$} represents the set of all the minimal solutions of \textbf{$S({\it A},{\it b})$}. Furthermore, we define $I(j)=\{ \, i\in I\, \, :\, \, a_{ij} >b_{i} \, \} $, \textbf{$\forall i\in I$}. According to [23], if \textbf{$S({\it A},{\it b})\ne \emptyset $}, maximum solution \textbf{$\bar{{\it x}}$} is easily obtained as stated in the following lemma. 

\noindent 
\vskip 0.2in
\noindent \textbf{Lemma 1. }Let $I(j)=\{ \, i\in I\, \, :\, \, a_{ij} >b_{i} \, \} $, \textbf{$\forall i\in I$}. if \textbf{$S({\it A},{\it b})\ne \emptyset $}, maximum solution \textbf{$\bar{{\it x}}$} is obtained as follows:
\begin{equation} \label{(2)} 
\bar{{\it x}}_{j} =\left\{\begin{array}{l} {{\mathop{\min }\limits_{i\in I(j)}} \left\{\, b_{i} \, \right\}\, \, \, \, \, \, \, I(j)\ne \emptyset } \\ {1\, \, \, \, \, \, \, \, \, \, \, \, \, \, \, \, \, \, \, \, \, \, \, \, I(j)=\emptyset } \end{array}\right. \, \, \, \, \, \, ,\forall j\in J 
\end{equation} 

\vskip 0.1in
\noindent \textbf{Definition 1. }Let \textbf{$\bar{{\it x}}$} be the maximum solution of \textbf{$S({\it A},{\it b})$}. For each $i\in I$, we define \textbf{$\bar{J}(i)=\{ \, j\in J\, \, :\, \, \min \, \{ \, a_{ij} ,\bar{{\it x}}_{j} \, \} =b_{i} \, \} $}. 

\noindent \textbf{}

\noindent \textbf{Definition 2.} Suppose that \textbf{$S({\it A},{\it b})\ne \emptyset $}. Let \textbf{$e:I\to \bigcup _{i\in I}\bar{J}(i) $} be a function defined on \textbf{$I$ }such that \textbf{$e\, (i)\in \bar{J}(i)$}, \textbf{$\forall i\in I$}. Also, let \textbf{$E$} denote the set of all the functions \textbf{$e$}. 

\noindent \textbf{}

\noindent \textbf{Remark 1. }From Definition 2, each \textbf{$e\in E$} can be equivalently considered as a set of ordered pairs \textbf{$\{ \, (\, 1\, ,\, j_{1} ),(\, 2\, ,\, j_{2} ),...,(\, m\, ,\, j_{m} )\, \} $} in witch \textbf{$j_{i} =e\, (i)$}, \textbf{$\forall i\in I$}. So, for the sake of simplicity, this set is usually represented only in terms of the second entries \textbf{$j_{1} $},\textbf{$j_{2} $},{\dots},\textbf{$j_{m} $}, and denoted by the notation \textbf{$e=[j_{1} ,j_{2} ,...,j_{m} ]$} as a finite sequence. 

\noindent \textbf{}

\noindent According to Definition 2, it is clear that \textbf{$\left|\, E\, \right|=\prod _{\, i\in I}\left|\, \bar{J}(i)\, \right| $}, where \textbf{$\left|\, E\, \right|$} and \textbf{$\left|\, \bar{J}(i)\, \right|$} (\textbf{$i=1,...,m$}) denote the cardinality of sets \textbf{$E$} and \textbf{$\bar{J}(i)$} (\textbf{$i=1,...,m$}), respectively.

\noindent 
\vskip 0.2in
\noindent \textbf{Definition 3. }Suppose that \textbf{$S({\it A},{\it b})\ne \emptyset $} and $e\in E$. Let \textbf{$I_{e} (j)=\{ \, i\in I\, \, :\, \, e\, (i)=j\, \} $}, \textbf{$\forall j\in J$}. Let \textbf{$\underline{{\it x}}\, (e)\in [0,1]^{n} $ }whose components are defined as follows:
\begin{equation} \label{(3)} 
\underline{{\it x}}\, (e)_{j} =\left\{\begin{array}{l} {{\mathop{\max }\limits_{i\in I_{e} (j)}} \left\{\, b_{i} \, \right\}\, \, \, \, \, \, \, I_{e} (j)\ne \emptyset } \\ {0\, \, \, \, \, \, \, \, \, \, \, \, \, \, \, \, \, \, \, \, \, \, \, \, I_{e} (j)=\emptyset } \end{array}\right. \, \, \, \, \, \, ,\forall j\in J 
\end{equation}

\noindent Based on the following theorem, each \textbf{$\underline{{\it x}}\, (e)$} ($e\in E$) is feasible to \textbf{$S({\it A},{\it b})$}. Moreover, although a solution \textbf{$\underline{{\it x}}\, (e)$} ($e\in E$) may not be minimal, each minimal solution \textbf{$\underline{{\it x}}\in \underline{S}({\it A},{\it b})$} can be written in the form of \textbf{$\underline{{\it x}}\, (e)$} for some $e\in E$ [17,23]. For this reason, each \textbf{$\underline{{\it x}}\, (e)$} ($e\in E$) is generally referred to as a \textbf{minimal candidate solution}.

\noindent 
\vskip 0.2in
\noindent \textbf{Theorem 1.  }Let \textbf{$S_{E} ({\it A},{\it b})=\{ \, \underline{{\it x}}\, (e)\, \, :\, \, e\in E\, \} $}. Then, \textbf{$\underline{S}({\it A},{\it b})\subseteq S_{E} ({\it A},{\it b})\subseteq S({\it A},{\it b})$}.

\noindent \textbf{}

\noindent Figure 1 depicts a typical feasible region \textbf{$S({\it A},{\it b})$} for problem (1) (shadowed region) in the 3-dimentional space \textbf{$S=[0,1]^{3} $}. As mentioned before, \textbf{$S({\it A},{\it b})$ }is often a non-convex set. Figure 1 also shows the different possible cases that may arise for minimal candidate solutions in the feasible region. As shown in the figure, \textbf{$\bar{{\it x}}$} is the maximum solution of \textbf{$S({\it A},{\it b})$}, and three minimal candidate solutions \textbf{$\underline{{\it x}}\, (e_{1} )$}, \textbf{$\underline{{\it x}}\, (e_{2} )$} and \textbf{$\underline{{\it x}}\, (e_{3} )$ }are both feasible and minimal. However, although the minimal candidate solution \textbf{$\underline{{\it x}}\, (e_{4} )$} is feasible to \textbf{$S({\it A},{\it b})$}, it is not a minimal solution, i.e., \textbf{$\underline{{\it x}}\, (e_{4} )\notin \underline{S}({\it A},{\it b})$}.  \textbf{}
\begin{center}
\noindent \includegraphics*[width=2.95in, height=2.25in, keepaspectratio=false]{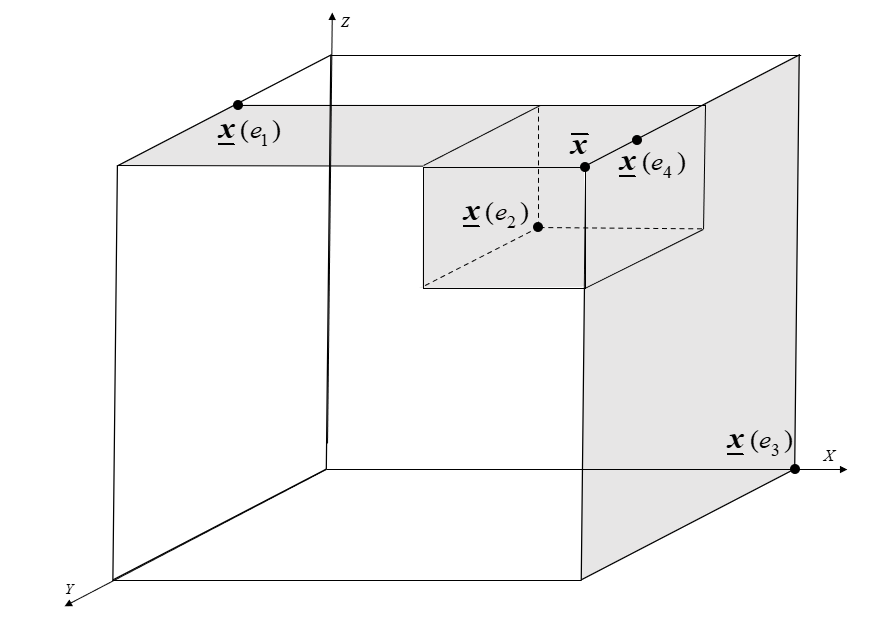}

\noindent \textbf{\footnotesize Figure1.} {\footnotesize The feasible region of problem(1) .}
\end{center}
\noindent \textbf{}

\noindent By the following theorem [7,17,23], the feasible region of problem(1) is completely determined by the unique maximum solution \textbf{$\bar{{\it x}}$} and a finite number of minimal candidate solutions \textbf{$\underline{{\it x}}\, (e)$} ($e\in E$). This theorem also shows the important role of the minimal candidate solutions in characterizing the feasible region of problem(1). Indeed, solutions \textbf{$\underline{{\it x}}\, (e)$} ($e\in E$) are the lower bounds of the feasible solution set for problem(1).

\noindent 
\vskip 0.1in
\noindent \textbf{Theorem 2. }If \textbf{$S({\it A},{\it b})\ne \emptyset $}, then \textbf{$S({\it A},{\it b})=\bigcup _{e\in E}[\underline{{\it x}}\, (e),\bar{{\it x}}] $}, where \textbf{$[\underline{{\it x}}\, (e),\bar{{\it x}}]$} is an \textbf{$n$}-dimensional interval including all points between \textbf{$\underline{{\it x}}\, (e)$} and \textbf{$\bar{{\it x}}$}.

\noindent 
\vskip 0.1in
\noindent \textbf{Definition 4 }[51]\textbf{. }If \textbf{$l_{j} ,u_{j} \in {\rm R}$} and \textbf{$l_{j} <u_{j} $} for \textbf{$j=1,...,k$}, the set of all points \textbf{${\it x}=[x_{1} ,...,x_{k} ]$} in \textbf{${\rm R}^{k} $} whose coordinates satisfy the inequalities \textbf{$l_{j} \le x_{j} \le u_{j} $} (\textbf{$1\le j\le k$}) is called a \textbf{$k$}--cell. Thus, a 1-cell is an interval such as \textbf{$[a,b]$}, a 2-cell is a rectangle, etc. Therefore, each \textbf{$k$}--cell  is  a closed convex set.
\vskip 0.2in
\noindent 

\noindent By Theorem 2, \textbf{$S({\it A},{\it b})$} is the set of all points between one maximum solution \textbf{$\bar{{\it x}}$}, and a finite number of minimal candidate solutions \textbf{$\underline{{\it x}}\, (e)$} (see Figure 1). In other words, \textbf{$S({\it A},{\it b})$} is a union of a finite number of closed convex cells \textbf{$[\underline{{\it x}}\, (e),\bar{{\it x}}]$ }($e\in E$). In Figure 1, \textbf{$S({\it A},{\it b})=\bigcup _{i=1}^{4}[\underline{{\it x}}\, (e_{i} ),\bar{{\it x}}] $ }where both \textbf{$[\underline{{\it x}}\, (e_{1} ),\bar{{\it x}}]$} and \textbf{$[\underline{{\it x}}\, (e_{3} ),\bar{{\it x}}]$} are 2-cells, \textbf{$[\underline{{\it x}}\, (e_{2} ),\bar{{\it x}}]$} is a 3-cell and \textbf{$[\underline{{\it x}}\, (e_{4} ),\bar{{\it x}}]$} is a 1-cell. As a consequence of Theorem 2, it turns out that \textbf{$S({\it A},{\it b})\ne \emptyset $ }iff \textbf{$\bar{{\it x}}\in S({\it A},{\it b})$}[7,17,23]. This corollary gives a simple necessary and sufficient condition for determining the feasibility of problem(1).

\noindent 
\vskip 0.2in
\noindent \textbf{Definition 5. }A matrix ${\it M}=(m_{ij} )_{m\, \times n} $ is said to be the minimal candidate matrix of set \textbf{$S({\it A},{\it b})$}, if 
\begin{equation} \label{(3)} 
m_{ij} =\left\{\begin{array}{l} {b_{i} \, \, \, \, \, \, \, j\in \bar{J}(i)} \\ {0\, \, \, \, \, \, \, \, j\notin \bar{J}(i)} \end{array}\right. \, \, \, \, \, \, ,\forall i\in I 
\end{equation} 
\vskip 0.1in
\noindent Using the minimal candidate matrix, another feasibility condition can be obtained as follows. \textbf{$S({\it A},{\it b})\ne \emptyset $} iff for each \textbf{$i\in I$}, there exists at least some \textbf{$j\in J$} such that \textbf{$m_{ij} >0$} [17].

\noindent Based on Remark 1 and Definitions 2 and 5, each \textbf{$e=[j_{1} ,j_{2} ,...,j_{m} ]$} in \textbf{$E$} can be interpreted as a path, denoted by \textbf{$(\, 1\, ,\, j_{1} )\to (\, 2\, ,\, j_{2} )\to ...\to (\, m\, ,\, j_{m} )$}, in the minimal candidate matrix ${\it M}$(as before, \textbf{$j_{i} =e\, (i)$}, \textbf{$\forall i\in I$}). In this respect, \textbf{$e$} (as a path in ${\it M}$) selects~one and only one~element \textbf{$(\, i\, ,\, j_{i} )$}~from~each row \textbf{$i$}~of matrix~${\it M}$. Also, it is important to note that each element or node \textbf{$(\, i\, ,\, j_{i} )$} included in \textbf{$e$} corresponds to a positive entry \textbf{$m_{ij_{i} } $} in ${\it M}$. Hereafter, each $e\in E$ is referred to as a \textbf{FRE-path}, that is, a path in the minimal candidate matrix. As a result, from Definition 3, each FRE-path $e\in E$ generates a minimal candidate solution \textbf{$\underline{{\it x}}\, (e)\in S_{E} ({\it A},{\it b})$}. So, each FRE-path $e$ can be associated to a closed convex cell \textbf{$[\underline{{\it x}}\, (e),\bar{{\it x}}]$} that is a subset of the feasible region \textbf{$S({\it A},{\it b})$} from Theorem 2. The following example illustrates the above-mentioned definitions.

\noindent 
\vskip 0.1in
\noindent \textbf{Example 1.} Consider the following FRE with the minimum t-norm $\varphi $ 
\[\left[\begin{array}{l} {{\rm 0.7\; \; \; 0.3\; \; \; 0.8\; \; \; 0.4\; \; \; 0.8\; \; \; 0.7}} \\ {{\rm 0.5\; \; \; 0.9\; \; \; 0.5\; \; \; 0.4\; \; \; 0.2\; \; \; 0.2}} \\ {{\rm 0.2\; \; \; 0.2\; \; \; 0.5\; \; \; 0.3\; \; \; 0.0\; \; \; 0.3}} \\ {{\rm 0\; \; \; \; \; \; 0.1\; \; \; 0\; \; \; \; \; \; 0.6\; \; \; 0.1\; \; \; 0}} \\ {{\rm 0.6\; \; \; 0.5\; \; \; 0.2\; \; \; 0.5\; \; \; 0.5\; \; \; 0.6}} \end{array}\right]\varphi {\it x=}\left[\begin{array}{l} {{\rm 0.7}} \\ {{\rm 0.5}} \\ {{\rm 0.3}} \\ {{\rm 0.1}} \\ {{\rm 0.6}} \end{array}\right]\] 
From Lemma 1, we have $I(1)=\emptyset $, $I(2)=\{ \, 2\, \} $, $I(3)=\{ \, 1,3\, \} $, $I(4)=\{ \, 4\, \} $, $I(5)=\{ \, 1\, \} $ and $I(6)=\emptyset $. Hence, maximum solution is obtained as \textbf{$\bar{{\it x}}=[{\rm 1\; ,0.5, 0.3,0.1 ,0.7 , 1}]$}. It is easy to verify that \textbf{$\bar{{\it x}}\in S({\it A},{\it b})$} which implies \textbf{$S({\it A},{\it b})\ne \emptyset $}. Also, from Definition 1, \textbf{$\bar{J}(1)=\{ \, 1,5,6\, \} $}, \textbf{$\bar{J}(2)=\{ \, 1,2\, \} $}, \textbf{$\bar{J}(3)=\{ \, 3,6\, \} $}, \textbf{$\bar{J}(4)=\{ \, 2,4,5\, \} $} and \textbf{$\bar{J}(5)=\{ \, 1,6\, \} $}. So, \textbf{$\left|\, E\, \right|=\left|\, \bar{J}(1)\, \right|\times \cdots \times \left|\, \bar{J}(5)\, \right|=72$}. Therefore, the number of all FRE-paths is equal to 72. Each FRE-path $e\in E$ generates a minimal candidate solution \textbf{$\underline{{\it x}}\, (e)$}. As an example, by selecting $e'=[{\rm 5,1,6,5,1}]$, from Definition 3 the corresponding solution is obtained as \textbf{$\underline{{\it x}}\, (e')=[{\rm 0.6\; ,\; 0\; ,\; 0\; ,\; 0\; ,\; 0.7\; ,\; 0.3}]$}. From Theorem 2, we know that \textbf{$[\underline{{\it x}}\, (e'),\bar{{\it x}}]\subseteq S({\it A},{\it b})$}, i.e., all points between \textbf{$\underline{{\it x}}\, (e')$} and \textbf{$\bar{{\it x}}$ }are feasible to problem(1). Moreover, the minimal candidate matrix ${\it M}=(m_{ij} )_{5\, \times 6} $ is obtained from Definition 5 as follows:
\begin{center}
\noindent \textbf{\includegraphics*[width=2.50in, height=1.51in, keepaspectratio=false]{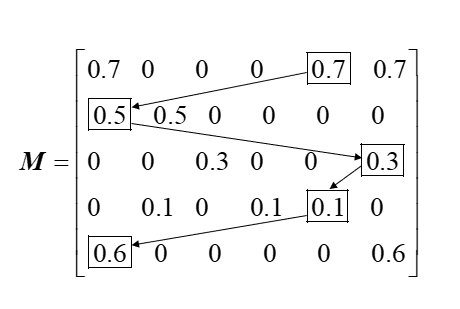}}
\end{center}
\noindent As mentioned before, $e'$ can be also denoted by \textbf{$(\, 1\, ,\, 5)\to (\, 2\, ,\, 1)\to (\, 3\, ,\, 6)\to (\, 4\, ,\, 5)\to (\, 5\, ,\, 1)$}. This FRE-path has been shown in matrix ${\it M}$. As is shown in the above matrix, each node included in FRE-path $e'$ (marked components) corresponds to a positive entry of ${\it M}$. Also, FRE-path $e'$ selects exactly one positive component from each row of ${\it M}$.  

\noindent 
\vskip 0.2in
\noindent \textbf{3. Phase (I) }
\vskip 0.1in
\noindent The original (discrete) ant colony optimization (ACO) algorithm [8] was proposed for solving combinatorial optimization problems such as scheduling, vehicle routing, timetabling, travelling salesman problem and so on. Many of these problems, especially those of practical relevance, are NP-hard. Generally, a model of a combinatorial optimization problem may be defined as follows [55]:

\noindent \textbf{}

\noindent \textbf{Definition 6. }A model \textbf{$P=({\it S},\Omega ,f)$} of a combinatorial optimization problem consists of a search space \textbf{${\it S}$} defined over a finite set of discrete decision variables \textbf{$X_{i} $} (\textbf{$i=1,...,m$}); a set \textbf{$\Omega $} of constraints among the variables; and an objective function \textbf{$f:{\it S}\to [0,\infty )$} to be minimized.  Furthermore, the generic variable \textbf{$X_{i} $}  takes values in \textbf{${\it D}_{i} =\{ v_{i}^{1} ,...,v_{i}^{\, \, \left|\, {\it D}_{i} \, \right|} \} $}. 

\noindent 
\vskip 0.2in
\noindent As mentioned in Section 1, FRE-ACO algorithm consists of two main phases. The purpose of Phase (I) is to find convex subset(s) of the feasible region, while Phase (II) produces better solutions by starting from feasible solutions selected from the convex feasible subsets. Each of the two phases is employed once during each iteration. In FRE-ACO, the pheromone is deposited on the positive entries of the minimal candidate matrix ${\it M}$. So, pheromone value \textbf{$\tau _{ij} $ }(associated with entry \textbf{$m_{ij} $}) is positive  if \textbf{$j\in \bar{J}(i)$} and \textbf{$\tau _{ij} =0$} otherwise. These values are summarized in the pheromone matrix ${\it \tau }=(\tau _{ij} )_{m\, \times n} $ . Pheromone matrix ${\it \tau }$ is updated at the end of each iteration, i.e., at the end of Phase (II). Moreover, based on a pheromone matrix ${\it \tau }$ in any arbitrary iteration, a probability matrix ${\it P}=(p_{ij} )_{m\, \times n} $ is obtained as follows
\begin{equation} \label{(4)} 
p_{ij} =\frac{\tau _{ij} }{\sum _{k=1}^{n}\tau _{ik}  }  
\end{equation} 
for each $i\in I$ and $j\in J$. According to Equation(4) , $p_{ij} $ is the probability of choosing $(i,j)$`th entry of ${\it M}$ (i.e., \textbf{$m_{ij} $}) from all entries in row $i$. In other words, for a fixed row $i$,  value $p_{ij} $ represents the probability of choosing column $j$ from $J=\{ 1,...,n\} $. At the beginning, the values $\tau _{ij} $ are set as follows:
\begin{equation} \label{(5)} 
\tau _{ij} =\left\{\begin{array}{l} {1\, \, \, \, \, \, \, j\in \bar{J}(i)} \\ {0\, \, \, \, \, \, j\notin \bar{J}(i)} \end{array}\right.  
\end{equation} 
Hence, from (4)  we obtain 
\begin{equation} \label{(6)}  
p_{ij} =\left\{\begin{array}{l} {{1\, \mathord{\left/ {\vphantom {1\,  \, \left|\, \bar{J}(i)\, \right|}} \right. \kern-\nulldelimiterspace} \, \left|\, \bar{J}(i)\, \right|} \, \, \, \, \, \, \, j\in \bar{J}(i)} \\ {0\, \, \, \, \, \, \, \, \, \, \, \, \, \, \, \, \, \, \, \, \, \, j\notin \bar{J}(i)} \end{array}\right.  
\end{equation} 

Given a probability matrix ${\it P}$, Phase (I) starts with $m_{1} $ ants. In this phase, each FRE-path is interpreted as a solution, and each artificial ant incrementally builds a partial FRE-path by selecting exactly one $j_{i} \in \bar{J}(i)$ for each row \textbf{$i$} in ${\it M}$. So, at the end of \textbf{$(i-1)$}`th step of the solution construction, a partial FRE-path \textbf{$[j_{1} ,...,j_{i-1} ]$} is obtained such that \textbf{$j_{s} \in \bar{J}(s)$} for \textbf{$s=1,...,i-1$}. At \textbf{$i$}`th step, ant $k$ visits \textbf{$i$}'th row  of ${\it M}$ and selects column $j_{i} \in \bar{J}(i)$\textbf{ } (equivalently, entry \textbf{$m_{ij_{i} } $ }in row \textbf{$i$ }of matrix ${\it M}$) with probability $p_{ij} $ as defined in (4). Each ant moves \textbf{$m$} steps to build an FRE-path \textbf{$e=[j_{1} ,...,j_{m} ]$}. At the end of Phase (I), with $m_{1} $FRE-paths $e_{1} $,{\dots},$e_{m_{\, 1} } $at hand we proceed to obtain the corresponding minimal candidate solutions \textbf{$\underline{{\it x}}\, (e_{1} )$},{\dots}, \textbf{$\underline{{\it x}}\, (e_{m_{\, 1} } )$} by Definition 3. Now, Theorem 2 implies the feasibility of $m_{1} $ convex cells \textbf{$[\underline{{\it x}}\, (e_{1} ),\bar{{\it x}}]$},{\dots}, \textbf{$[\underline{{\it x}}\, (e_{m_{\, 1} } ),\bar{{\it x}}]$}. So, in order to access the feasible cells \textbf{$[\underline{{\it x}}\, (e_{l} ),\bar{{\it x}}]$}, $l=1,..,m_{1} $, it is sufficient to know their lower bounds (i.e., \textbf{$\underline{{\it x}}\, (e_{l} )$}, $l=1,..,m_{1} $); because, the upper bound of each feasible cell is the maximum solution \textbf{$\bar{{\it x}}$} that can be always found by (2) at the beginning. The preceding discussions can be summarized as an algorithm shown below:

\noindent 
\vskip 0.2in
\noindent \textbf{Algorithm 1} (Phase(I)).
\vskip 0.1in

\noindent 1. Get $m_{1} $and pheromone matrix ${\it \tau }$.

\noindent 2. Generate the probability matrix ${\it P}$ by (4).

\noindent 3. For $l=1$ to $m_{1} $ 

\noindent 4.        For $i=1$ to $m$ 

\noindent 5.                select $j_{i} $ with the probability $p_{ij} $ stated by (4). 

\noindent 6.                set \textbf{$e_{l} (i)=j_{i} $}. 

\noindent 7.         End For

\noindent 8.         Set \textbf{$e_{l} =[j_{1} ,...,j_{m} ]$}.

\noindent 9.         By (3), compute minimal candidate solution \textbf{$\underline{{\it x}}\, (e_{l} )$} generated by the FRE-path \textbf{$e_{l} $}.

\noindent 10. End For

\noindent \textbf{   }

\noindent It is to be noted that by setting \textbf{${\it S}=E$}, \textbf{$X_{i} =e\, (i)$} and \textbf{${\it D}_{i} =\bar{J}(i)$}, the problem of finding FRE-paths \textbf{$e\in E$} can be represented as a combinatorial problem \textbf{$P=({\it S},\Omega ,f)$} as defined in Definition 6. The only difference is that Phase (I) considers no objective function to be optimized, because its purpose is to find some feasible subsets in the feasible region of problem (1). \textbf{}

\noindent 
\vskip 0.2in
\noindent \textbf{4. Phase (II)}
\vskip 0.2in
\noindent In [55], the authors proposed a way to effectively apply ACO to continuous optimization problems following the fundamental framework of original ACO. The fundamental idea underlying ACO${}_{R}$ is substituting the discrete probability distributions used in ACO algorithms for combinatorial problems with probability density functions in the solution construction step. This section describes Phase (II) of FRE-ACO that is based on ACO${}_{R}$${}_{\ \ }$algorithm.

\noindent 
\vskip 0.2in
\noindent 4.1. Initialization 
\vskip 0.2in
\noindent Many optimization methods randomly generate the initial population. This strategy works well when dealing with unconstrained optimization problems. However, for a constrained optimization problem, randomly generated solutions may not be feasible. In FRE-ACO, the initial population (with population size \textbf{$S_{pop} $}) is given by randomly generating the solutions inside $m_{1} $ feasible cells \textbf{$[\underline{{\it x}}\, (e_{l} ),\bar{{\it x}}]$} (one solution per cell). These cells are easily accessible by considering the lower bounds \textbf{$\underline{{\it x}}\, (e_{l} )$} obtained at the end of Phase (I) and the maximum solution \textbf{$\bar{{\it x}}$}.  Each solution \textbf{$s_{\, l} $} in Phase (II) consists of three parts and it may be denoted by \textbf{$s_{\, l} =[s_{\, l} ({\it x})\, ,\, s_{\, l} (lb)\, ,\, s_{\, l} (e)]$}; \textbf{$s_{\, l} ({\it x})$} is an \textbf{$n$}-dimensional point randomly selected from \textbf{$[\underline{{\it x}}\, (e_{l} ),\bar{{\it x}}]$}, i.e., \textbf{$s_{\, l} ({\it x})\in [\underline{{\it x}}\, (e_{l} ),\bar{{\it x}}]$}, \textbf{$s_{\, l} (lb)=\underline{{\it x}}\, (e_{l} )$} that shows the lower bound of the cell including the feasible point \textbf{$s_{\, l} ({\it x})$}, and \textbf{$s_{\, l} (e)=e_{l} $ } that enables us to access the  FRE-path by which the cell \textbf{$[\underline{{\it x}}\, (e_{l} ),\bar{{\it x}}]$} is generated. In the following, a typical feasible region \textbf{$S({\it A},{\it b})$} has been depicted in the 2-dimentional space \textbf{$S=[0,1]^{2} $}. Figure 2 also shows cell \textbf{$[\underline{{\it x}}\, (e_{l} ),\bar{{\it x}}]$} (shadowed region) and a new randomly generated point \textbf{$s_{\, l} ({\it x})$} in \textbf{$[\underline{{\it x}}\, (e_{l} ),\bar{{\it x}}]$}. 

\noindent 
\begin{center}
\noindent \includegraphics*[width=2.51in, height=1.60in, keepaspectratio=false]{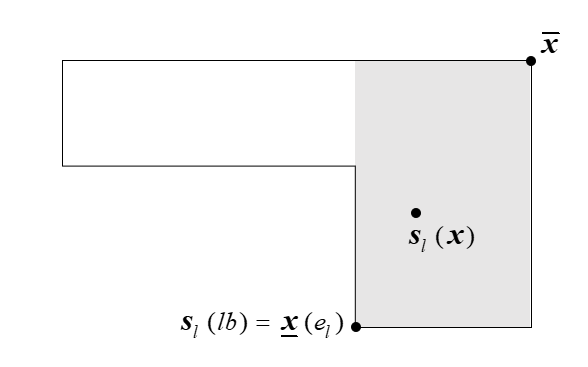}

\noindent \textbf{\footnotesize Figure2.}{\footnotesize Initialization step.}
\end{center}
\noindent 
\vskip 0.2in
\noindent The algorithm for generating the initial solutions is simply obtained as follows:
\vskip 0.2in
\noindent \textbf{Algorithm 2} (Initialization).
\vskip 0.1in
\noindent 1. Get maximum solution \textbf{$\bar{{\it x}}$}, $m_{1} $lower bounds \textbf{$\underline{{\it x}}\, (e_{1} )$} ,{\dots},\textbf{$\underline{{\it x}}\, (e_{m_{\, 1} } )$}and objective function \textbf{$f({\it x})$}.

\noindent 2. For $l=1$ to $m_{1} $ 

\noindent 3.        Select a random \textbf{$n$}-dimensional solution point \textbf{$s_{\, l} ({\it x})$} from \textbf{$[\underline{{\it x}}\, (e_{l} ),\bar{{\it x}}]$}. 

\noindent 4.         Set \textbf{$s_{\, l} (lb)=\underline{{\it x}}\, (e_{l} )$} and \textbf{$s_{\, l} (e)=e_{l} $}. 

\noindent 5.         Evaluate and add the solution \textbf{$s_{\, l} =[s_{\, l} ({\it x})\, ,\, s_{\, l} (lb)\, ,\, s_{\, l} (e)]$} to the archive $T$.

\noindent 6. End For

\noindent 7.  Sort solutions \textbf{$s_{\, l} $} , \textbf{$l=1,...,m_{\, 1} $}, in an ascending order of objective values \textbf{$f(s_{\, l} ({\it x}))$}.
\vskip 0.2in
\noindent 

\noindent For future reference, we give an example to illustrate Algorithm 2.

\noindent \textbf{}

\noindent \textbf{Example 2.} Consider the maximum solution \textbf{$\bar{{\it x}}=[{\rm 1\; ,\; 0.5\; ,\; 0.3\; ,\; 0.1\; ,\; 0.7\; ,\; 1}]$}, the FRE-path $e'=[{\rm 5,1,6,5,1}]$ and its corresponding minimal candidate solution \textbf{$\underline{{\it x}}\, (e')=[{\rm 0.6\; ,\; 0\; ,\; 0\; ,\; 0\; ,\; 0.7\; ,\; 0.3}]$} stated in Example 1. Suppose  that $e'$ has been already generated by some ant $l\in \{ \, 1,...,m_{\, 1} \, \} $ in Phase (I). Therefore, we have \textbf{$e_{l} =e'$} and \textbf{$\underline{{\it x}}\, (e_{l} )=\underline{{\it x}}\, (e')$}. Consequently, in order to generate a solution \textbf{$s_{\, l} =[s_{\, l} ({\it x})\, ,\, s_{\, l} ({\it lb})\, ,\, s_{\, l} (e)]$}, Algorithm 2 chooses a random point from the cell \textbf{$[\underline{{\it x}}\, (e_{l} )\, ,\, \bar{{\it x}}]$}, e.g., \textbf{$s_{\, l} ({\it x})=[0.8{\rm \; ,\; 0.3\; ,\; 0.2\; ,\; 0\; ,\; 0.7\; ,\; 1}]$}. Moreover, by assuming that \textbf{$f({\it x})=x_{1} x_{4} -x_{2} x_{3} x_{5} +x_{6}^{2} $} is the objective function of problem (1), solution \textbf{$s_{\, l} $} is evaluated by \textbf{$f(s_{\, l} ({\it x}))={\rm 0.958}$}.

\noindent 
\vskip 0.2in
\noindent 4.2. Sampling process
\vskip 0.2in
\noindent In Phase (II), the algorithm stores a set of \textbf{$S_{pop} $} so-far-best solutions in a solution archive $T$. The solution archive is used to create a probability distribution of promising solutions over the continuous search space. Initially, $T$ is filled with solutions \textbf{$s_{\, l} =[s_{\, l} ({\it x})\, ,\, s_{\, l} (lb)\, ,\, s_{\, l} (e)]$}, \textbf{$l=1,...,S_{pop} $}, that are randomly generated by Algorithm 2. The \textbf{$j$}`th variable of the \textbf{$l$}`th solution point \textbf{$s_{\, l} ({\it x})$} is denoted by \textbf{$s_{\, l} ({\it x})_{j} $}. Also, the quality of each solution \textbf{$s_{\, l} $} is evaluated by \textbf{$f(s_{\, l} ({\it x}))$}, that is, the objective function value of its corresponding feasible point. Phase (II) iteratively refines the solution archive by generating $m_{2} $ new solutions and then keeping only the best \textbf{$S_{pop} $} solutions of the $S_{pop} +m_{2} $ solutions that are available. The \textbf{$S_{pop} $} solutions \textbf{$s_{\, l} $} in $T$ are always sorted according to their quality from the best to worst, that is, solution \textbf{$s_{\, l} $} has rank $l$.  Therefore, the first solution, \textbf{$s_{\, 1} $}, is the best one in the sense that it includes the best solution point \textbf{$s_{\, 1} ({\it x})$} (stored in its first part) with the smallest objective value \textbf{$f(s_{\, 1} ({\it x}))$}. The weight of the archive solution \textbf{$s_{\, l} $} is calculated by the following Gaussian function:
\begin{equation} \label{GrindEQ__8_} 
\omega _{\, l} =\frac{1}{\sqrt{2\pi } \, q\, S_{pop} } e^{-\, \frac{1}{2} \, \, \left(\frac{l\, -\, 1}{\, q\, \, S_{pop} } \right)^{2} } ,\, \, \, \, \, \, l=1,...,S_{pop}  
\end{equation} 
which essentially defines the weight to be a value of the Gaussian function associated with the ranking of solution \textbf{$l$} in the archive, mean $1$, and standard deviation \textbf{$q\, S_{pop} $}, where \textbf{$q$} is a parameter of the algorithm. When \textbf{$q$} is small, the best-ranked solutions are strongly preferred, and when it is large, the probability becomes more uniform. 

\noindent During the solution generation process, each coordinate is treated independently.  The generation of new solutions is accomplished by a sampling process as follows. First, an archive solution \textbf{$s_{\, l} $} is chosen with the following probability:
\begin{equation} \label{GrindEQ__9_} 
p_{l} =\frac{\omega _{\, l} }{\sum _{\, k=1}^{\, S_{pop} }\omega _{\, k}  }  
\end{equation} 
Then, for \textbf{$j\in J=\{ \, 1\, ,\, ...\, ,\, n\, \} $}, the algorithm samples around the \textbf{$j$}`th component of the solution point of the selected solution, i.e., \textbf{$s_{l}^{j} $}, using a Gaussian function \textbf{$g_{l}^{j} $}, where \textbf{$l$} is a solution index and \textbf{$j$} is a coordinate index. The Gaussian function for coordinate \textbf{$j$} is given by:
\begin{equation} \label{GrindEQ__10_} 
g_{l}^{j} ({\it x})=\frac{1}{\sigma _{l}^{j} \sqrt{2\pi } } e^{-\, \frac{1}{2} \, \, \left(\frac{x\, -\, \mu \, _{l}^{j} }{\sigma _{l}^{j} } \right)^{2} }  
\end{equation} 
\textbf{ }with the mean \textbf{$\mu _{\, l}^{\, j} =s_{\, l} ({\it x})_{j} $}, and the standard deviation \textbf{$\sigma _{l}^{\, j} $} equal to
\begin{equation} \label{GrindEQ__11_} 
\sigma _{l}^{\, j} =\xi \, \sum _{\, k=1}^{\, S_{pop} }\frac{\left|\, s_{\, k} ({\it x})_{j} -s_{\, l} ({\it x})_{j} \, \right|}{S_{pop} -1}  \, \, \, \, \, \, \, \, \, \, \, \, \, \, \, ,\, \, \, l=1,...,S_{pop}  
\end{equation} 
which is the average distance between the \textbf{$j$}`th variable of the feasible point \textbf{$s_{\, l} ({\it x})$} and the \textbf{$j$}`th variable of the other feasible points in the archive. The parameter \textbf{$\xi >0$} has an effect similar to that of the pheromone evaporation rate in ACO. The higher the value of \textbf{$\xi $}, the lower the convergence speed of the algorithm. By repeating(10) for \textbf{$j\in \{ \, 1\, ,\, ...\, ,\, n\, \} $}, we obtain a solution \textbf{${\it x}'$} around \textbf{$s_{\, l} ({\it x})$}. Since \textbf{${\it x}'$} may be infeasible, we adjust its components to guarantee that both \textbf{${\it x}'$} and \textbf{$s_{\, l} ({\it x})$} will be in the same cell, i.e., \textbf{$[\underline{{\it x}}\, (e_{l} )\, ,\, \bar{{\it x}}]$}. To this end, the algorithm uses the maximum solution \textbf{$\bar{{\it x}}$} and the lower bound \textbf{$\underline{{\it x}}\, (e_{l} )$} stored in the second part of \textbf{$s_{\, l} ({\it x})$}, i.e., \textbf{$s_{\, l} ({\it lb})=\underline{{\it x}}\, (e_{l} )$}. So, this is accomplished by the following assignment: 
\begin{equation} \label{GrindEQ__12_} 
{\it x}'_{j} \leftarrow \min \, \{ \, \max \, \{ \, {\it x}'_{j} \, ,\, \underline{{\it x}}\, (e_{l} )_{j} \, \} \, ,\, \bar{{\it x}}_{j} \, \} \, \, \, \, \, \, ,\, j=1\, ,\, ...\, ,\, n 
\end{equation} 
Finally, since each solution archive contains three parts, we define the new solution \textbf{$s'$} by \textbf{$s'=[s'({\it x}),s'({\it lb}),s'(e)]$} where \textbf{$s'({\it x})={\it x}'$}, \textbf{$s'({\it lb})=s_{\, l} ({\it lb})=\underline{{\it x}}\, (e_{l} )$} and \textbf{$s'(e)=s_{\, l} (e)=e_{l} $}. The solution generation process is repeated $m_{2} $ times and the new solutions \textbf{$s'_{1} $},{\dots},\textbf{$s'_{m_{\, 2} } $} are added to the archive $T$. Each new solution that is added to $T$ is evaluated and ranked. The sampling process is presented as the following algorithm:

\noindent 
\vskip 0.2in
\noindent \textbf{Algorithm 3} (Sampling process).
\vskip 0.1in
\noindent 1. Get $q$, $\xi $, $m_{2} $, \textbf{$f({\it x})$} and sorted solutions \textbf{$s_{\, l} =[s_{\, l} ({\it x})\, ,\, s_{\, l} ({\it lb})\, ,\, s_{\, l} (e)]$}, \textbf{$l=1\, ,\, ...\, ,\, S_{pop} $}.

\noindent 2. For $k=1$ to $m_{2} $ 

\noindent 3.        According to (9), select an archive solution \textbf{$s_{\, l} $} with probability \textbf{$p_{l} $}.

\noindent 4.        According to (10) and (11), sample Gaussian function \textbf{$g_{l}^{j} ({\it x})$} for $j=1\, ,\, ...\, ,\, n$.

\noindent 5.        Generate the modified point \textbf{${\it x}'$} by assignment (12).

\noindent 6.        Generate new solution \textbf{$s'_{k} =[s'_{k} ({\it x})\, ,\, s'_{k} ({\it lb})\, ,\, s'_{k} (e)]$} by setting \textbf{$s'_{k} ({\it x})={\it x}'$}, \textbf{$s'_{k} ({\it lb})=\underline{{\it x}}\, (e_{l} )$}

\noindent            \textbf{ }and \textbf{$s'_{k} (e)=e_{l} $}.

\noindent 7. End For

\noindent 8. Evaluate and add the new solutions \textbf{$s'_{k} $}, \textbf{$k=1\, ,\, ...\, ,\, m_{2} $}, to the archive $T$.

\noindent 9. Update the archive by keeping the \textbf{$S_{pop} $} best solutions and removing the $m_{2} $ worst ones. 

\noindent 
\vskip 0.2in
\noindent 4.3. Updating pheromone matrix 
\vskip 0.2in
\noindent At the end of Phase (II), the solution archive is also used to update pheromone matrix ${\it \tau }$. For this purpose, each archive solution \textbf{$s_{\, l} =[s_{\, l} ({\it x})\, ,\, s_{\, l} ({\it lb})\, ,\, s_{\, l} (e)]$} updates the matrix ${\it \tau }$ as follows:
\begin{equation} \label{GrindEQ__13_} 
\tau _{i,e_{l} (i)} \, =\, \, \tau _{i,e_{l} (i)} +Q\, e^{-\, f(s_{\, l} ({\it x}))} \, \, \, \, \, \, ,\forall i\in I 
\end{equation} 
where $e_{l} =s_{\, l} (e)$ (stored in the third part of \textbf{$s_{\, l} $}), \textbf{$\tau _{i,e_{l} (i)} $} denotes the entry in the \textbf{$i$}`th row and \textbf{$e_{l} (i)$}`th column of the matrix ${\it \tau }$, \textbf{$f(s_{\, l} ({\it x}))$} is the objective value of the feasible point \textbf{$s_{\, l} ({\it x})$} (stored in the first part of \textbf{$s_{\, l} $}), \textbf{$Q$} is a positive adjusting constant and \textbf{$Q\, e^{-\, f(s_{\, l} ({\it x}))} $} is the quality of pheromone laid on each entry $(i,e_{l} (i))$ of ${\it \tau }$, \textbf{$\forall i\in I$}. The pheromone matrix ${\it \tau }$ is updated \textbf{$S_{pop} $} times by Equation (13). Finally, it is modified by the evaporation rate \textbf{$\rho \in [0,1)$ }as follows:
\begin{equation} \label{GrindEQ__14_} 
{\it \tau }=\, \, (1-\rho )\, {\it \tau } 
\end{equation} 
In the archive $T$, better solutions \textbf{$s_{\, l} $} include better feasible points \textbf{$s_{\, l} ({\it x})$} with less objective values \textbf{$f(s_{\, l} ({\it x}))$} that imply higher values of \textbf{$Q\, e^{-\, f(s_{\, l} ({\it x}))} $}. Therefore, better solutions deposit higher amounts of pheromone on the components of matrix ${\it \tau }$. Moreover, according to (13), each archive solution \textbf{$s_{\, l} $} increases the probability of choosing its corresponding FRE-path (i.e., \textbf{$s_{\, l} (e)=e_{l} $}) by Phase (I) at the next iteration. This strategy guides Phase (I) towards those FRE-paths that generate promising cells, that is, those cells that includes better solution points. 

\noindent \textbf{}

\noindent The algorithm for updating the pheromone matrix is described as follows:

\noindent \textbf{}

\noindent \textbf{Algorithm 4} (Updating pheromone matrix).
\vskip 0.1in
\noindent 1. Get $\rho $, $Q$, \textbf{$f({\it x})$}, \textbf{$s_{\, l} =[s_{\, l} ({\it x})\, ,\, s_{\, l} ({\it lb})\, ,\, s_{\, l} (e)]$}, \textbf{$l=1\, ,\, ...\, ,\, S_{pop} $}, and the pheromone matrix ${\it \tau }$.

\noindent 2. For $l=1$ to $S_{pop} $ 

\noindent 3.        Update the pheromone matrix ${\it \tau }$ by (13).

\noindent 4. End For

\noindent 5. Modify the pheromone matrix ${\it \tau }$ by (14).

\noindent \textbf{}

\noindent \textbf{5. FRE-ACO algorithm and a comparative study}
\vskip 0.2in

\noindent Based on Algorithms 1-4, The different steps of FRE-ACO algorithm is expressed as follows. As is described in Algorithm 5, at the first iteration (\textbf{$t=1$}), Phase (I) starts with \textbf{$m_{\, 1} =S_{pop} $} ants to generate \textbf{$S_{pop} $} FRE-paths \textbf{$e_{l} $} and their corresponding minimal candidate solutions \textbf{$\underline{{\it x}}\, (e_{l} )$}(\textbf{$l=1\, ,\, ...\, ,S_{pop} $}). Then, by using the \textbf{$S_{pop} $} cells \textbf{$[\underline{{\it x}}\, (e_{l} ),\bar{{\it x}}]$}, Algorithm 2 creates the initial archive \textbf{$T$} with the archive size \textbf{$S_{pop} $}. At the first iteration, Phase (II) does not perform the sampling process, and therefore the initial archive is directly utilized to update the pheromone matrix \textbf{$\tau $}. In contrast, at the next iterations, Phase (I) uses only one ant (\textbf{$m_{\, 1} =1$}) to find just one FRE-path \textbf{$e_{l} $} based on the biased (by pheromone) probabilistic choice of the entries \textbf{$\tau _{ij} $}. Consequently, except the first iteration, Phase (I) produces only one minimal candidate solution \textbf{$\underline{{\it x}}\, (e_{l} )$}. Subsequently, using the feasible cell \textbf{$[\underline{{\it x}}\, (e_{l} ),\bar{{\it x}}]$}, Algorithm 2 generates one new solution and adds the solution to the archive \textbf{$T$}. This strategy avoids a~premature convergence~in FRE-ACO Algorithm and increases the ability of the algorithm in escaping from local optima in a promising cell. Moreover, Phase (II) performs the sampling process two times (\textbf{$m_{\, 2} =2$}) at each iteration \textbf{$t$} such that  \textbf{$t>1$}. To sum up, Phase (I) provides feasible convex cells for Phase (II) to explore in proper regions for generating an initial archive. Additionally, Phase (I) also participates in the exploitation step by constructing the FRE-paths that generates promising cells. On the other hand, Phase (II) improves the solutions by sampling around the better solution points in the archive, and by increasing the probability of choosing the FRE-paths that generate the promising cells. Meanwhile, Phase (II) also participates in the exploration step by finding a random solution point in a promising cell, and by adding this solution point to the archive.

\noindent \textbf{}

\noindent \textbf{Algorithm 5} (FRE-ACO algorithm).
\vskip 0.1in
\noindent 1. Get fuzzy matrix \textbf{${\it A}$}, fuzzy vector \textbf{${\it b}$}, objective function \textbf{$f({\it x})$}, population size $S_{pop} $, constants $q$, $\xi $, $\rho $and $Q$, and the maximum number of iterations \textbf{$t_{max} $}. 

\noindent 2. Compute the maximum solution \textbf{$\bar{{\it x}}$} by (2). 

\noindent 3. If \textbf{$\bar{{\it x}}\notin S({\it A},{\it b})$}, then stop (problem is infeasible).

\noindent 4. Generate the initial pheromone matrix ${\it \tau }$ by (5). 

\noindent \textbf{\qquad     Phase (I):}

\noindent 5. \textbf{ }Employ Algorithm 1 with \textbf{$m_{\, 1} =S_{pop} $} to generate \textbf{$S_{pop} $} FRE-paths.

\noindent \textbf{\qquad     Phase (II):}

\noindent 6. Create the initial archive $T$ by Algorithm 2.

\noindent 7. Update the pheromone matrix ${\it \tau }$ by Algorithm 4.

\noindent 8. For $t=2$ to $t_{\max } $ 

\noindent            \textbf{\qquad Phase (I): }

\noindent 9.\textbf{        }Employ Algorithm 1 with \textbf{$m_{\, 1} =1$}  to generate a promising FRE-path.\textbf{}

\noindent \textbf{\qquad           Phase (II):}

\noindent 10.      Employ Algorithm 2 to generate a new solution and add the solution to the archive.

\noindent 11.      Employ the sampling process with \textbf{$m_{\, 2} =2$} by Algorithm 3. 

\noindent 12.      Update the pheromone matrix ${\it \tau }$ by employing Algorithm 4.

\noindent 13. End For

\noindent 
\vskip 0.2in
\noindent   The following points highlight some of the main characteristics of the proposed algorithm and some important differences between FRE-ACO and the original ACO and ACO${}_{R}$ algorithms. 

\noindent 
\begin{itemize}
\item \noindent FRE-ACO algorithm consists of two phases that are based on discrete and continuous ant colony optimization algorithms. Employing these two ideas in the same algorithm provides a suitable tradeoff between exploration and exploitation.

\item \noindent  Unlike the general-purpose original ACO algorithms, FRE-ACO algorithm has been designed specifically for solving nonlinear optimization problems with fuzzy relation constraints. For this purpose, the current algorithm solves the problem by taking advantage of the structure of the solutions set that is a non-convex set.

\item \noindent  In FRE-ACO ( Phase (I)), the purpose of ants is to find FRE-paths, i.e., the paths in the minimal candidate matrix ${\it M}$. 

\item \noindent  In ACO, the~pheromone value on~a path is updated after the path construction while in FRE-ACO algorithm FRE-paths are not evaluated in Phase (I) where the FRE-paths are built. Indeed, the quality of an FRE-path is evaluated based on the quality of a continuous solution (generated in Phase (II)) in the cell generated by the FRE-path.  

\item \noindent  In FRE-ACO algorithm, each archive solution includes three parts; first, a feasible solution point; second, the lower bound of the cell containing the solution point; and third, the FRE-path that generates the lower bound.

\item \noindent  In the original ACO${}_{R}$, the best archive solutions have the highest chances to be selected in the sampling process. However, in FRE-ACO algorithm, the best solutions increase the probability of choosing the FRE-paths that generate their corresponding cells. More precisely, if \textbf{$s_{\, 1} =[s_{\, 1} ({\it x})\, ,\, s_{\, 1} ({\it lb})\, ,\, s_{\, 1} (e)]$} is the best archive solution at \textbf{$t$}`th iteration and its FRE-path (i.e., \textbf{$s_{\, 1} (e)$}) is built in Phase (I) at \textbf{$(t+1)$}'th iteration, then Algorithm 2 adds a new solution \textbf{$s_{\, l} =[s_{\, l} ({\it x})\, ,\, s_{\, l} ({\it lb})\, ,\, s_{\, l} (e)]$} to the archive such that \textbf{$s_{\, l} ({\it lb})=s_{\, 1} ({\it lb})$}, \textbf{$s_{\, l} (e)=s_{\, 1} (e)$}, and \textbf{$s_{\, l} ({\it x})$} and \textbf{$s_{\, 1} ({\it x})$} are in the same cell \textbf{$[s_{\, 1} ({\it lb}),\bar{{\it x}}]$}. Only in the case that \textbf{$f(s_{\, 1} ({\it x}))\le f(s_{\, l} ({\it x}))$}, solution \textbf{$s_{\, 1} $} can maintain its rank in the archive with the highest probability to be selected in the sampling process.
\end{itemize}
\noindent  
\vskip 0.2in
\noindent Moreover, to see how the current algorithm is situated comparing the other algorithms designed for FRE nonlinear optimization problems, we compare theoretically our algorithm with the genetic algorithms presented in [15,34]. In addition, an experimental comparison is given in the next section.

\noindent In [15,34], the selection strategy is only based on the rank of the solutions in the population. But in FRE-ACO algorithm, this strategy depends on the rank of the archive solutions as well as the probability of choosing their corresponding FRE-paths and the quality of the random solution that is added to the archive at each iteration \textbf{$t>1$}.  

\noindent In [34], the proposed mutation operator decreases one variable $x_{j} $ of point ${\it x}$\textbf{ }to a random number between $[0\, ,\, x_{j} )$\textbf{ }each time. In this mutation operator, a decreasing variable often followed by increasing several other variables to guarantee the feasibility of a new solution. This operator was improved in [15] by finding a proper variable, and by decreasing it to zero. However, in the proposed algorithm, new generated points are always selected from the closed convex cells that are the subsets of the feasible region. 

\noindent Lu and Fang [34] used the ``three-point'' crossover operator defined by three points (two parents ${\it x}_{1} $ and ${\it x}_{2} $, and the maximum solution \textbf{$\bar{{\it x}}$}) and two operators called ``contraction'' and ``extraction''. Both contraction and extraction operators are employed between ${\it x}_{1} $\textbf{ }and ${\it x}_{2} $, and between ${\it x}_{i} $ ($i=1,2$) and \textbf{$\bar{{\it x}}$}. However, from the four mentioned cases, only one case certainly results in a feasible offspring (i.e., the contraction between ${\it x}_{i} $ ($i=1,2$) and \textbf{$\bar{{\it x}}$}). Therefore, for the other three cases, the feasibility of the new generated solutions must be checked by substituting them into the fuzzy relational equations as well as the constraints $x_{j} \in [0,1]$, $\forall j\in J$. In [15], the authors introduced a crossover operator using only one parent each time and for producing two feasible offsprings ${\it x}_{new1} $ and ${\it x}_{new2} $. Offspring ${\it x}_{new1} $\textbf{ }is obtained as a random point on the line segment between ${\it x}'$\textbf{ }(generated by the mutation operator)\textbf{ }and \textbf{$\bar{{\it x}}$}. But, offspring  ${\it x}_{new2} $\textbf{ }lies close to its parent. In the current algorithm, the new point \textbf{$s'({\it x})$} (the first part of \textbf{$s'$}) is generated by sampling around \textbf{$s_{\, l} ({\it x})$} (the first part of the archive solution \textbf{$s_{\, l} $}) in such a way that both \textbf{$s'({\it x})$} and \textbf{$s_{\, l} ({\it x})$} are in the same feasible cell \textbf{$[\underline{{\it x}}\, (e_{l} )\, ,\, \bar{{\it x}}]$}. This cell is easily accessible by considering the lower bound \textbf{$\underline{{\it x}}\, (e_{l} )$} that is stored in the second part of \textbf{$s_{\, l} $}, i.e., \textbf{$s_{\, l} ({\it lb})=\underline{{\it x}}\, (e_{l} )$}, and the maximum solution \textbf{$\bar{{\it x}}$} that is obtained easily at the beginning of the algorithm.

\noindent In [15,34], initial population is generated by selecting random solutions from the cell \textbf{$F=[\underline{{\it x}}\, ,\, \bar{{\it x}}]$}, where \textbf{$\underline{{\it x}}$} is obtained by the maximization on all the elements of set \textbf{$S=\{ \, \underline{{\it x}}\, (e)\, \, :\, \, e\in E\, \, ,\, \, \underline{{\it x}}\, (e)\le \, \bar{{\it x}}\, \} $}. The preference of this strategy is mainly based on the finding of region \textbf{$F$} without calculating the minimal solutions. nevertheless, \textbf{$\underline{{\it x}}$} has high sensitivity to each component of each minimal solution. In other words, if any component of any minimal point is large, then the same component of \textbf{$\underline{{\it x}}$} will be large that restricts region \textbf{$F$}. Region \textbf{$F$} may be even smaller when $S$includes some solutions that are not minimal (e.g., point \textbf{$\underline{{\it x}}\, (e_{4} )$ }in Figure 1) . In Figure 1, \textbf{$F$} is obtained as the line segment between \textbf{$\underline{{\it x}}\, (e_{4} )$} and \textbf{$\bar{{\it x}}$}. Consequently, \textbf{$F$} is usually a small region that decreases the diversity of individuals in an initial population.   In contrast, the proposed algorithm uses cells \textbf{$[\underline{{\it x}}\, (e_{l} ),\bar{{\it x}}]$} in which the solutions \textbf{$\underline{{\it x}}\, (e_{l} )$} are the lower bounds for the feasible solutions (Theorem 2). This strategy increases the ability of the algorithm to expand the search space for finding new solutions. 

\noindent \textbf{}

\noindent \textbf{6. Experimental results}
\vskip 0.2in
\noindent In this section, we present the experimental results for evaluating the performance of our algorithm. As mentioned before, we can make a comparison between the FRE-ACO algorithm and the genetic algorithms proposed in [15,34] for solving nonlinear optimization problems subjected to fuzzy relational equations. For this purpose, we apply these algorithms to 10 test problems described in Appendix. The test problems have been randomly generated in different sizes by the procedure given in [15]. Table 1 shows the optimal objective values for the ten test problems where \textbf{$\# $ }denotes the problem number. To perform a fair comparison, we follow the same experimental setup for the parameters $\theta =0.5$, $\xi =0.01$, $\lambda =0.995$and $\gamma =1.005$ as suggested in [34], and $q=0.1$ as suggested by the authors in [15]. Also, we consider \textbf{$S_{pop} =50$ }for all the three algorithms. The summary of the parameters we used for FRE-ACO is presented in Table 2. Moreover, according to Algorithm 5, the number of ants used by Phase (I) at the \textbf{$t$}'th  iteration is \textbf{$m_{\, 1} =S_{pop} $} for \textbf{$t=1$}, and \textbf{$m_{\, 1} =1$} for \textbf{$t>1$}. Also, the number of sampling processes employed by Phase (II) at the \textbf{$t$}'th iteration is \textbf{$m_{\, 2} =0$} for \textbf{$t=1$}, and \textbf{$m_{\, 2} =2$} for \textbf{$t>1$}. Finally, 30 experiments are performed for all the algorithms and for each test problem, that is, each of the preceding algorithm is executed 30 times for each test problem. Also, the maximum number of iterations is equal to 100 for all the methods.

\noindent 

\noindent 
\vskip 0.1in
\begin{tabular}{p{2.5in}p{3.5in}} 
 \noindent \textbf{\footnotesize Table 1 }

\noindent {\footnotesize The optimal values of the test problems.}

\noindent \begin{tabular}{p{0.2in}|p{1.2in}} \hline 
\# & Optimal values \\ \hline 
1 & - 0.0096019 \\  
2 & 0.8197 \\ 
3 & 80.3752 \\ 
4 & - 0.39657 \\ 
5 & - 0.27162 \\ 
6 & 1.2612 \\  
7 & 140.4693 \\  
8 & - 0.10108 \\  
9 & 1.277 \\  
10 & 55.7954 \\ \hline
\end{tabular}
&
\textbf{\footnotesize Table 2}

  {\footnotesize Summary of the parameters used by FRE-ACO.}

 \begin{tabular}{p{1.5in}|p{0.6in}|p{0.6in}} \hline 
Parameter & Symbol & Value \\ 
speed of convergence & \textbf{$\xi $} & 1 \\ 
Locality of the search process & \textbf{$q$} & 0.0125 \\ 
Evaporation rate & \textbf{$\rho $} & 0.5 \\ 
Adjusting constant & \textbf{$Q$} & 1 \\ 
Archive size & \textbf{$S_{pop} $} & 50 \\ \hline 
\end{tabular}
\end{tabular}
\vskip 0.1in

\noindent The computational results of the 10 test problems are shown in Table 3, where Avg, Mdn and SD denote the average best-so-far, median best-so-far and standard deviation, respectively. In all the cases, the results that are marked with ``*'' indicate the better cases. In Table 3, the results have been averaged over 30 runs and the average best-so-far solution and median of the best solution in the last iterations are reported. This table also includes the best results, \textbf{$f_{best} $}, found by the algorithms over 30 runs. Table 3 demonstrates the ability of the FRE-ACO to detect the optimal solutions of problem (1). According to the results in this table, the best objective values computed by the FRE-ACO and the optimal objective values (see Table 1) match very well. Moreover, Table 3 shows that the current algorithm finds the optimal values faster than those obtained by the other related algorithms and hence has a higher convergence rate, even for the same solutions. The only exceptions are test problems 3 and 4 where all the results are the same for FRE-ACO and the GA presented by Ghodousian et. al. [15]. Also, the good convergence rate of the proposed algorithm could be concluded from Figures 3-12. Furthermore, the proposed algorithm has the smallest standard deviation (good stability) for all the test problems when compared against the other algorithms. Consequently, FRE-ACO could be a robust global search algorithm for nonlinear FRE optimization problems even though the feasible domain is generally non-convex. 

\noindent 
\vskip 0.1in
\noindent \textbf{\footnotesize Table 3}

\noindent {\footnotesize A comparison between the results found by the GA proposed by Lu and Fang, the GA proposed by Ghodousian and Babalhavaeji and FRE-ACO algorithms for the test problems. The results have been averaged over 30 runs. Maximum number of iterations=100.}

\noindent\begin{tabular}{p{0.7in}|p{0.4in}|p{1.1in}|p{1.5in}|p{0.7in}} \hline 
\#  &  & GA(Lu and Fang) & GA(Ghodousian et. al.) & FRE-ACO \\  \hline 
1 & Avg\textbf{} & - 0.0075 & - 0.0041 & - 0.0096${}^{*}$ \\ 
 & Mdn\textbf{} & - 0.0089 & - 0.0096${}^{*}$ & - 0.0096${}^{*}$ \\ 
 & Sd\textbf{} & 0.0236 & 0.0250 & 0.0017 \\ 
 & \textbf{$f_{best} $} & - 0.0096${}^{*}$ & - 0.0096${}^{*}$ & - 0.0096${}^{*}$ \\ \hline 
2 & Avg\textbf{} & 0.8258 & 0.8206 & 0.8197${}^{*}$ \\ 
 & Mdn\textbf{} & 0.8212 & 0.8201 & 0.8197${}^{*}$ \\ 
 & Sd\textbf{} & 0.2444 & 0.1689 & 0.0388 \\  
 & \textbf{$f_{best} $} & 0.8197${}^{*}$ & 0.81971 & 0.8197${}^{*}$ \\ \hline 
3 & Avg\textbf{} & 80.3776 & 80.3752${}^{*}$ & 80.3752${}^{*}$ \\ 
 & Mdn\textbf{} & 80.3752 & 80.3752${}^{*}$ & 80.3752${}^{*}$ \\  
 & Sd\textbf{} & 0.3067 & 0.2474 & 0.0283 \\ 
 & \textbf{$f_{best} $} & 80.3752${}^{*}$ & 80.3752${}^{*}$ & 80.3752${}^{*}$ \\  \hline 
4 & Avg\textbf{} & - 0.3933 & - 0.3966${}^{*}$ & - 0.3966${}^{*}$ \\ 
 & Mdn\textbf{} & - 0.3957 & - 0.3966${}^{*}$ & - 0.3966${}^{*}$ \\ 
 & Sd\textbf{} & 0.0142 & 0.0133 & 0.0021 \\ 
 & \textbf{$f_{best} $} & - 0.3966${}^{*}$ & - 0.3966${}^{*}$ & - 0.3966${}^{*}$ \\  \hline 
5 & Avg\textbf{} & - 0.2411 & - 0.2444 & - 0.2716${}^{*}$ \\ 
 & Mdn\textbf{} & - 0.2432 & - 0.2716${}^{*}$ & - 0.2716${}^{*}$ \\ 
 & Sd\textbf{} & 0.1008 & 0.1267 & 0.0188 \\ 
 & \textbf{$f_{best} $} & - 0.2715 & - 0.2716${}^{*}$ & - 0.2716${}^{*}$ \\ \hline 
6 & Avg\textbf{} & 1.4492 & 1.3766 & 1.3656${}^{*}$ \\  
 & Mdn\textbf{} & 1.443 & 1.2612${}^{*}$ & 1.2612${}^{*}$ \\ 
 & Sd\textbf{} & 0.2652 & 0.2627 & 0.1938 \\ 
 & \textbf{$f_{best} $} & 1.272 & 1.2612${}^{*}$ & 1.2612${}^{*}$ \\ \hline 
7 & Avg\textbf{} & 140.6311 & 140.5467 & 140.4705${}^{*}$ \\ 
 & Mdn\textbf{} & 140.5874 & 140.5267 & 140.4699${}^{*}$ \\  
 & Sd\textbf{} & 0.8632 & 0.8382 & 0.3094 \\  
 & \textbf{$f_{best} $} & 140.4694 & 140.4722 & 140.4693${}^{*}$ \\ \hline
8 & Avg\textbf{} & - 0.0849 & - 0.1005 & - 0.101${}^{*}$ \\ 
 & Mdn\textbf{} & - 0.0850 & - 0.101${}^{*}$ & - 0.101${}^{*}$ \\ 
 & Sd\textbf{} & 0.0147 & 0.0134 & 0.0044 \\  
 & \textbf{$f_{best} $} & - 0.0995 & - 0.101${}^{*}$ & - 0.101${}^{*}$ \\ \hline
9 & Avg\textbf{} & 1.2882 & 1.3028 & 1.2813${}^{*}$ \\ 
 & Mdn\textbf{} & 1.2806 & 1.277${}^{*}$ & 1.277${}^{*}$ \\ 
 & Sd\textbf{} & 0.0523 & 0.0555 & 0.0239 \\  
 & \textbf{$f_{best} $} & 1.277${}^{*}$ & 1.277${}^{*}$ & 1.277${}^{*}$ \\  \hline 
10 & Avg\textbf{} & 56.9274 & 56.1322 & 55.8338${}^{*}$ \\ 
 & Mdn\textbf{} & 56.8232 & 56.0079 & 55.7957${}^{*}$ \\ 
 & Sd\textbf{} & 10.7861 & 10.0182 & 0.6052 \\  
 & \textbf{$f_{best} $} & 55.8051 & 55.8166 & 55.7954${}^{*}$ \\ \hline 
\end{tabular}

\vskip 0.2in

\noindent Table 4 contains the average errors of the solutions found by the methods. The results have been averaged over 3000 iterations (over all the iterations). The last row of Table 4 provides the mean square error over all the test problems. Based on the table, we can note which method has the lower error on average. The results, in Table 4, demonstrate that FRE-ACO produces high quality solutions with low average error (good accuracy) compared with the solutions obtained by the other methods. 

\noindent 
\vskip 0.2in
\noindent \textbf{\footnotesize Table 4}

\noindent {\footnotesize  Errors averaged over 3000 iterations (obtained from 30 runs, each of which contains 100 iterations).}

\noindent\begin{tabular}{p{0.6in}|p{1.1in}|p{1.5in}|p{0.7in}} \hline 
\# & GA(Lu and Fang) & GA(Ghodousian et. al.) & FRE-ACO \\ \hline 
1 & 0.0141 & 0.0094 & 0.0002 \\  
2 & 0.1054 & 0.0366 & 0.0063 \\  
3 & 0.0882 & 0.0257 & 0.0024 \\ 
4 & 0.0089 & 0.0025 & 0.0002 \\  
5 & 0.0883 & 0.0473 & 0.0034 \\ 
6 & 0.3361 & 0.1692 & 0.1159 \\ 
7 & 0.5888 & 0.4755 & 0.1062 \\  
8 & 0.0236 & 0.0049 & 0.0011 \\ 
9 & 0.0321 & 0.0276 & 0.0050 \\  
10 & 5.2138 & 2.9753 & 0.2768 \\ 
MSE & 2.7672 & 0.9112 & 0.0101 \\ \hline 
\end{tabular}
\vskip 0.2in

\noindent The number of function evaluations required in the execution of each method is reported in Table 5. The results have been averaged over 30 runs. As mentioned in Section 5, Lu and Fang's GA may produce a different number of new solutions by the crossover operator at an iteration. So, in contrast to the other algorithms, Lu and Fang's GA uses a different number of function evaluations for solving the test problems. As is shown in this table, FRE-ACO requires the least number of function evaluations for all the test problems compared to the other considered algorithms. 

\noindent 
\vskip 0.2in
\noindent \textbf{\footnotesize Table 5}

\noindent {\footnotesize Number of Function Evaluations averaged over 30 runs.}

\noindent\begin{tabular}{p{0.7in}|p{1.1in}|p{1.5in}|p{0.7in}} \hline 
\# & GA (Lu \& Fang) & GA (Ghodousian et al.) & FRE-ACO \\ \hline 
1 & 351.2333 & 350 & 347 \\ 
2 & 350.9667 & 350 & 347 \\ \
3 & 350.9333 & 350 & 347 \\  
4 & 351.0333 & 350 & 347 \\ 
5 & 351.1 & 350 & 347 \\ 
6 & 351.1 & 350 & 347 \\ 
7 & 350.9 & 350 & 347 \\ 
8 & 350.9667 & 350 & 347 \\  
9 & 351.2 & 350 & 347 \\ 
10 & 351.1 & 350 & 347 \\ \hline 
\end{tabular}
\vskip 0.2in

\noindent The main advantage of FRE-ACO is remarkable convergence rate, while it is competitive in terms of quality too. Particularly, FRE-ACO significantly outperforms the mentioned genetic algorithms in term of convergence rate and is better in most cases in terms of quality or is equal in the worst case. Also, by considering the notion of the complexity, suppose that algorithms A and B have the same output value (e.g., convergence rate) by performing k and 2k function evaluations, respectively. So, although A and B have the same output, the performance of algorithm A is reasonably two times better than that of algorithm B. Therefore, by considering the output values with respect to the number of function evaluations, we can make a fair comparison between the performances of the algorithms [55]. Table 6 shows the results of Related-Samples Wilcoxon Signed Rank Test to illustrate significant difference between the algorithms. 
\vskip 0.2in

\noindent 

\noindent \textbf{\footnotesize Table 6}

\noindent {\footnotesize RELATED TEST WILCOXON- CRITERIA ALPHA=0.05  CILEVEL=95}

\noindent\begin{tabular}{p{1.7in}|p{1.5in}|p{1.5in}} \hline 
\   & Difference between FRE-ACO and GA (Lu \& Fang) & Difference between FRE-ACO and GA (Ghodousian et al.)  \\ \hline 
  Mean of Negative Ranks & 5.50 & 4.50 \\ 
  Mean of Positive Ranks & 0.00 & 0.00 \\ 
  Sum of Negative Ranks & 55.00 & 36.00 \\  
  Sum of Positive Ranks & 0.00 & 0.00 \\ 
  Avg & 158.2201 & 98.3256 \\ 
  SD & 256.4982 & 162.0671 \\ 
  SEM & 81.1118 & 51.2501 \\ 
  Z  & 2.803 & 2.521 \\  
  p-value  & .005 & .012 \\ \hline 

\end{tabular}
\vskip 0.2in

\noindent According to the results, the negative and positive ranks of the difference of both groups of variables and the sum of the negative and positive marked ranks can be seen in the table. 
The number, average and sum of negative ranks are more than positive ranks, it can be concluded that the results obtained from the FRE-ACO are smaller than GA (Lu \& Fang) and GA (Ghodousian et al.).
The asymptotic probability value for the test for the hypothesis that there is no difference between GA (Lu \& Fang) and FRE-ACO, it is equal to 0.005.
Also, the asymptotic probability value for the test for the hypothesis that there is no difference between the GA (Ghodousian et al.) and FRE-ACO, is equal to 0.012
 According to the following relationships, the null hypothesis is rejected. This means that there is a significant difference between the algorithms.

\noindent 1.	Sig. ={$ 0.005 <0.05 $}

\noindent 2.	Sig. ={$ 0.012 < 0.05 $}

\noindent The Z with 2.803 and 2.521 values at 5 percent significance level also shows this differences.

\noindent 

\noindent 
\begin{center}
\noindent \includegraphics*[width=3.26in, height=2.59in, keepaspectratio=false, trim=0.19in 0.04in 0.27in 0.15in]{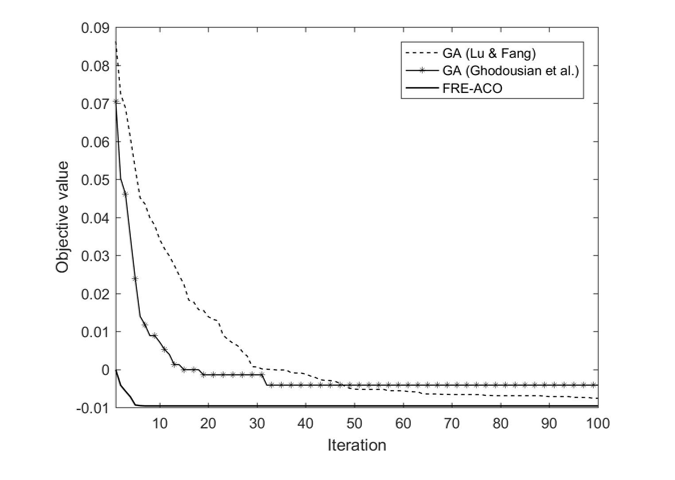}\textbf{}

\noindent \textbf{\footnotesize Figure 3.} {\footnotesize The performance of the algorithms on test problem 1.}
\end{center}
\vskip 0.2in
\noindent 
\begin{center}
\noindent \includegraphics*[width=3.25in, height=2.59in, keepaspectratio=false, trim=0.19in 0.04in 0.27in 0.16in]{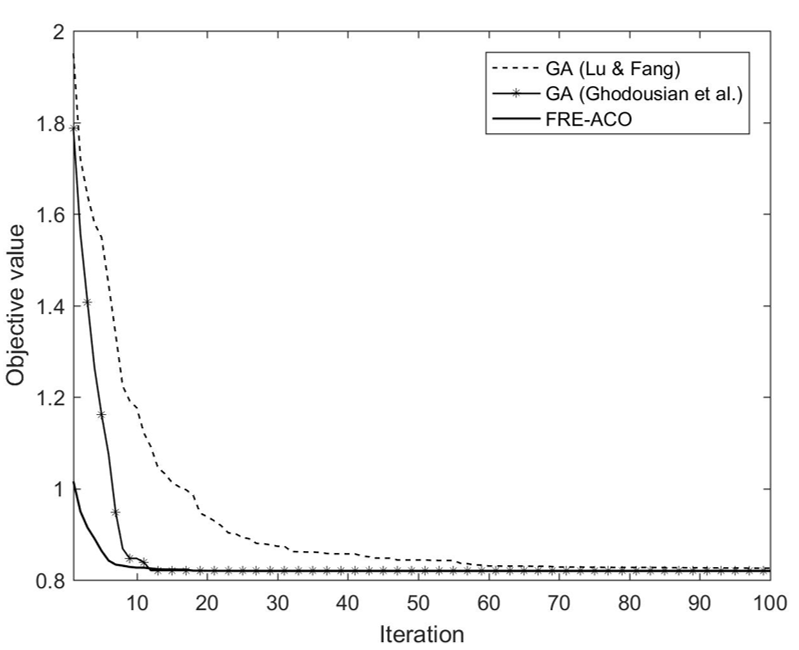}\textbf{}

\noindent \textbf{\footnotesize Figure 4.} {\footnotesize The performance of the algorithms on test problem 2.}
\end{center}
\vskip 0.4in
\noindent 
\begin{center}
\noindent \includegraphics*[width=3.27in, height=2.56in, keepaspectratio=false, trim=0.14in 0.04in 0.28in 0.17in]{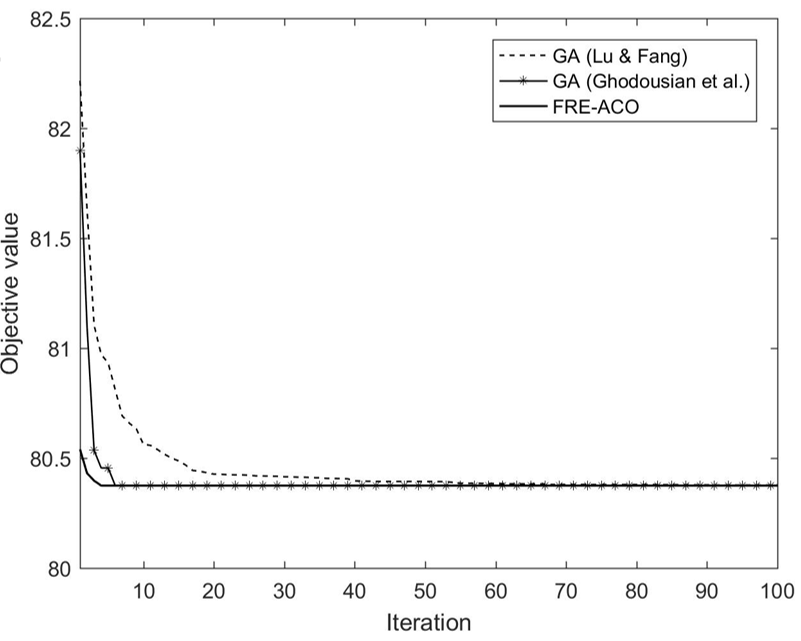}\textbf{}

\noindent \textbf{\footnotesize Figure 5.} {\footnotesize The performance of the algorithms on test problem 3.}
\end{center}
\vskip 0.2in
\noindent 
\begin{center}
\noindent \includegraphics*[width=3.36in, height=2.62in, keepaspectratio=false, trim=0.12in 0.04in 0.28in 0.16in]{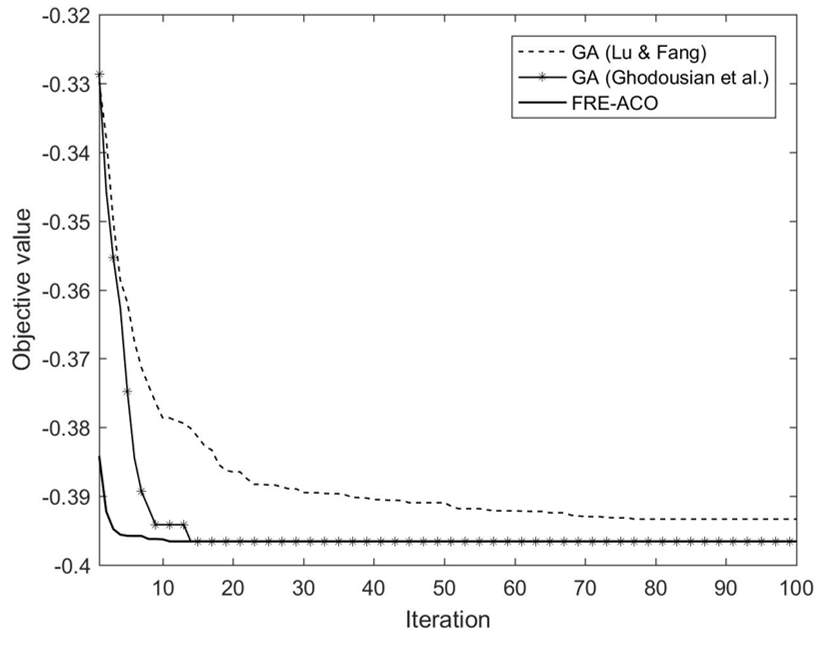}\textbf{}

\noindent \textbf{\footnotesize Figure 6.} {\footnotesize The performance of the algorithms on test problem 4.}
\end{center}
\vskip 0.4in
\noindent 
\begin{center}
\noindent \includegraphics*[width=3.36in, height=2.61in, keepaspectratio=false, trim=0.12in 0.04in 0.27in 0.17in]{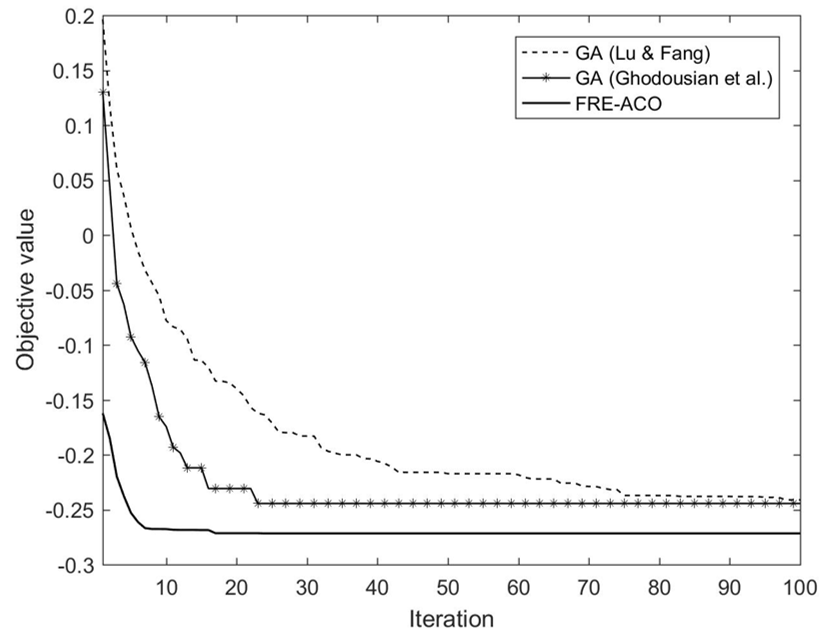}\textbf{}

\noindent \textbf{\footnotesize Figure 7.} {\footnotesize The performance of the algorithms on test problem 5.}
\end{center}
\vskip 0.2in
\noindent 

\noindent 
\begin{center}
\noindent \includegraphics*[width=3.21in, height=2.56in, keepaspectratio=false, trim=0.20in 0.04in 0.28in 0.17in]{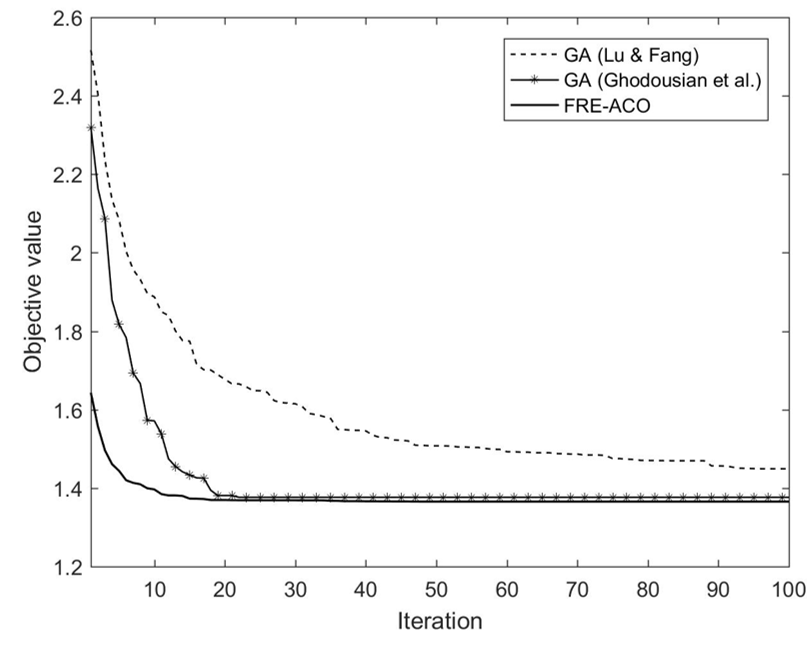}\textbf{}

\noindent \textbf{\footnotesize Figure 8.} {\footnotesize The performance of the algorithms on test problem 6.}
\vskip 0.4in
\noindent 

\noindent \includegraphics*[width=3.38in, height=2.61in, keepaspectratio=false, trim=0.10in 0.04in 0.28in 0.17in]{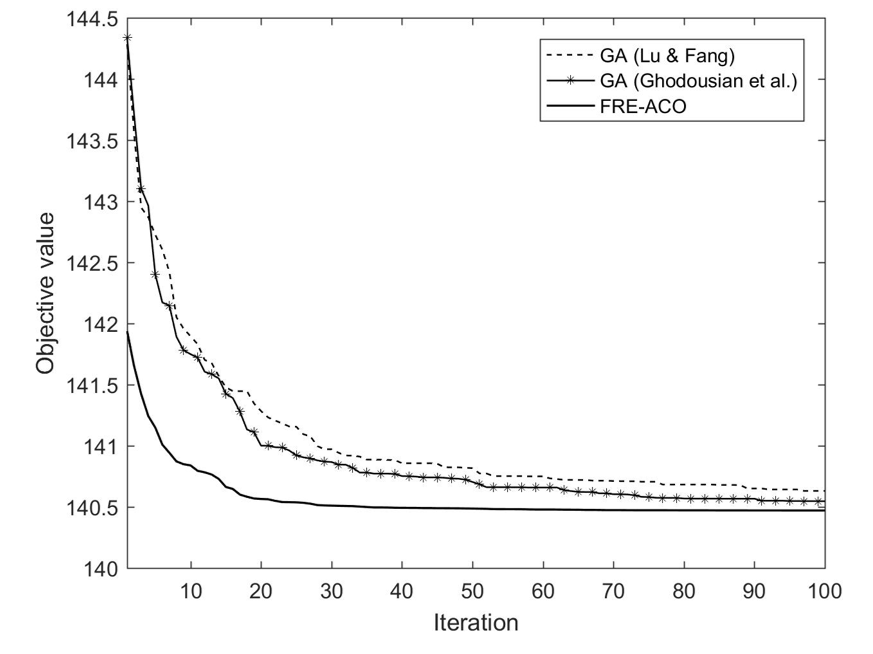}\textbf{}

\noindent \textbf{\footnotesize Figure 9.} {\footnotesize The performance of the algorithms on test problem 7.}
\vskip 0.2in
\noindent 

\noindent \includegraphics*[width=3.38in, height=2.64in, keepaspectratio=false, trim=0.12in 0.04in 0.28in 0.18in]{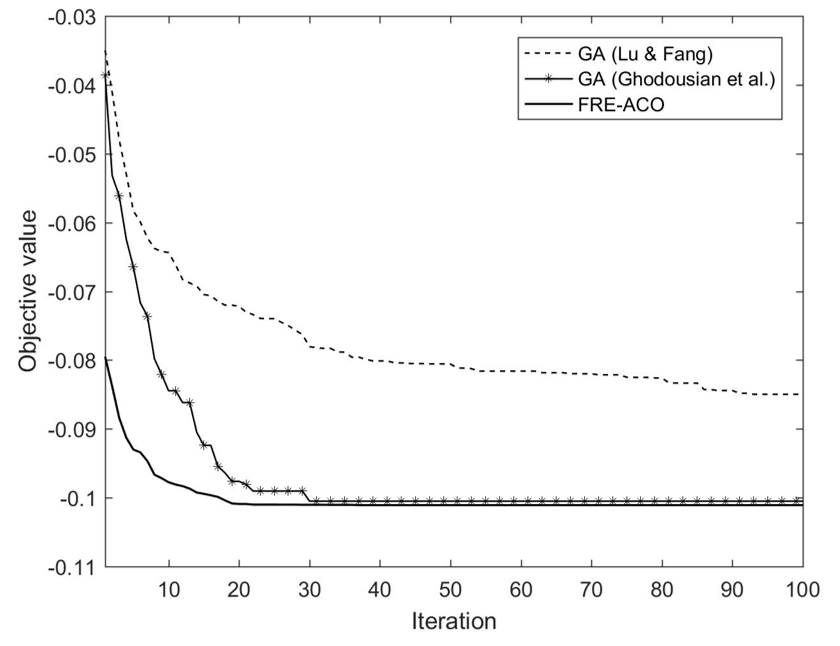}\textbf{}

\noindent \textbf{\footnotesize Figure 10.} {\footnotesize The performance of the algorithms on test problem 8.}
\vskip 0.4in
\noindent 

\noindent \includegraphics*[width=3.28in, height=2.56in, keepaspectratio=false, trim=0.14in 0.04in 0.27in 0.17in]{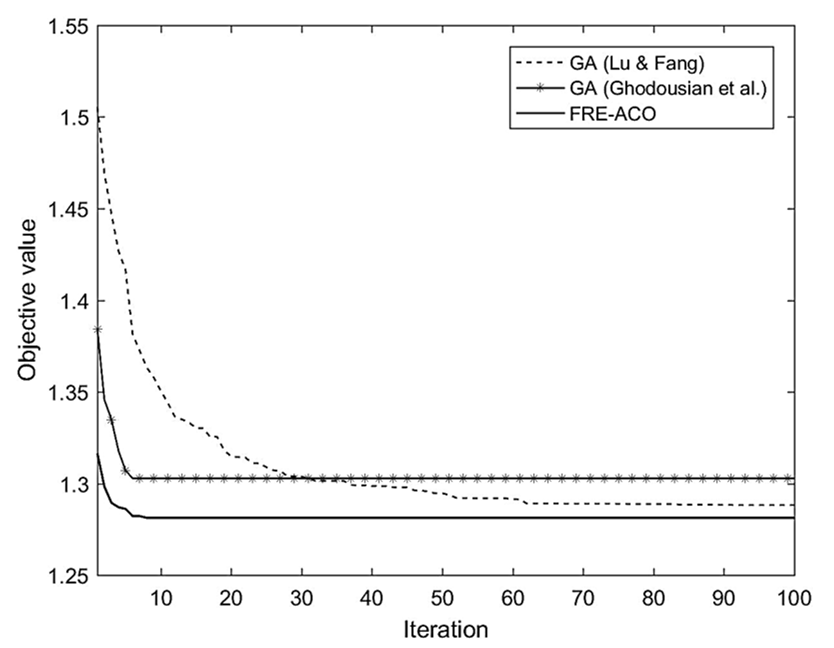}\textbf{}

\noindent \textbf{\footnotesize Figure 11.} {\footnotesize The performance of the algorithms on test problem 9.}
\vskip 0.2in
\noindent 

\noindent \includegraphics*[width=3.32in, height=2.61in, keepaspectratio=false, trim=0.17in 0.04in 0.28in 0.18in]{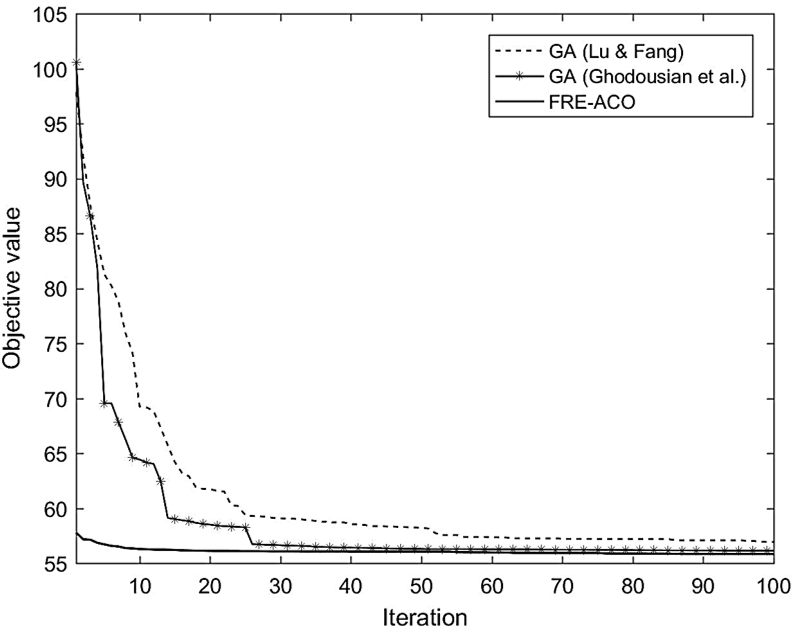}\textbf{}

\noindent \textbf{\footnotesize Figure 12.} {\footnotesize The performance of the algorithms on test problem 10.}
\end{center}
\noindent 
\vskip 0.2in
\noindent \textbf{6. Conclusion}
\vskip 0.2in
\noindent In this paper, an ACO-based algorithm (FRE-ACO) was proposed for solving nonlinear optimization problems with fuzzy relational equations defined by the minimum t-norm. In the proposed algorithm, the idea of both discrete and continuous ant colony optimization algorithms were used to simultaneously tackle the resolution of the feasible region (that is an NP-hard combinatorial problem) and the optimization of nonlinear functions subjected to non-convex regions. Experiments were performed by the proposed method for the ten randomly generated feasible test problems. We conclude that FRE-ACO can find the optimal solutions for all the cases with a great convergence rate. Moreover, a comparison was made between FRE-ACO algorithm and two genetic algorithms (GA), Lu and Fang's GA and Ghodousian and Babalhavaeji's GA, which solve the nonlinear optimization problems subjected to the FREs defined by max-min composition. The results showed that the proposed method finds better solutions (or the same solutions with a higher convergence rate) compared with the solutions obtained by the other algorithms. Furthermore, FRE-ACO produces high quality solutions with a less average error (good accuracy) and less standard deviation (good stability) for all the test problems. Moreover, the results demonstrated that FRE-ACO requires the least number of function evaluations when compared against the considered genetic algorithms.As the future work, we investigate other heuristic algorithms to solve the nonlinear optimization problems restricted to fuzzy relational equations.

\noindent   
\vskip 0.2in
\noindent \textbf{Appendix }
\vskip 0.2in
\noindent \textbf{Test Problem 1}
\[f({\it x})=Ln\, (0.5+x_{1}^{2} x_{2} +x_{3} )-x_{4}^{2} +x_{5} x_{6} \] 
\[{\it A}=\left[\begin{array}{l} {{\rm 0.1795\; \; \; 0.9194\; \; \; 0.6636\; \; \; 0.2774\; \; \; 0.1598\; \; \; 0.8240}} \\ {{\rm 0.6077\; \; \; 0.4035\; \; \; 0.7116\; \; \; 0.7843\; \; \; 0.2370\; \; \; 0.6523}} \\ {{\rm 0.9365\; \; \; 0.3045\; \; \; 0.2486\; \; \; 0.3578\; \; \; 0.7174\; \; \; 0.5950}} \\ {{\rm 0.1200\; \; \; 0.1041\; \; \; 0.0306\; \; \; 0.1692\; \; \; 0.0409\; \; \; 0.2391}} \end{array}\right]\] 
\[{\it b}^{T} =[{\rm 0.8501\; ,\; 0.5064\; ,\; 0.3178\; ,\; 0.1263}]\] 
\vskip 0.2in
\textbf{Test Problem 2}
\[f({\it x})=\sin \, (x_{1} x_{2} )+(1-\cos \, (x_{1} x_{3} ))+x_{4} +x_{5}^{2} +x_{6}^{3} \] 
\[{\it A}=\left[\begin{array}{l} {{\rm 0.7515\; \; \; 0.0275\; \; \; 0.8476\; \; \; 0.9867\; \; \; 0.9788\; \; \; 0.8664}} \\ {{\rm 0.8268\; \; \; 0.1254\; \; \; 0.8037\; \; \; 0.5647\; \; \; 0.9216\; \; \; 0.0806}} \\ {{\rm 0.3641\; \; \; 0.5033\; \; \; 0.5515\; \; \; 0.5515\; \; \; 0.1354\; \; \; 0.0200}} \\ {{\rm 0.0826\; \; \; 0.5188\; \; \; 0.2551\; \; \; 0.1630\; \; \; 0.1430\; \; \; 0.1882}} \\ {{\rm 0.6395\; \; \; 0.7118\; \; \; 0.2717\; \; \; 0.3305\; \; \; 0.2062\; \; \; 0.0968}} \\ {{\rm 0.8624\; \; \; 0.0310\; \; \; 0.0027\; \; \; 0.0939\; \; \; 0.0637\; \; \; 0.1416}} \end{array}\right]\] 
\[{\it b}^{T} =[{\rm 0.8629\; ,\; 0.8629\; ,\; 0.5515\; ,\; 0.3789\; ,\; 0.3789\; ,\; 0.1846}]\] 
\vskip 0.2in
\textbf{Test Problem 3}
\[f({\it x})=(x_{1} +10x_{2} )^{2} +5(x_{3} -x_{4} )^{2} +(x_{2} -2x_{3} )^{4} +10(x_{1} -x_{4} )^{4} -x_{5} -x_{6} +(2x_{7} +x_{8} )^{2} \] 
\[{\it A}=\left[\begin{array}{l} {{\rm 0.7715\; \; \; 0.1104\; \; \; 0.8930\; \; \; 0.2939\; \; \; 0.9973\; \; \; 0.8488\; \; \; 0.7017\; \; \; 0.7097}} \\ {{\rm 0.7837\; \; \; 0.9825\; \; \; 0.6753\; \; \; 0.7218\; \; \; 0.3572\; \; \; 0.1096\; \; \; 0.0955\; \; \; 0.2023}} \\ {{\rm 0.2266\; \; \; 0.5948\; \; \; 0.9205\; \; \; 0.1939\; \; \; 0.0579\; \; \; 0.2825\; \; \; 0.8445\; \; \; 0.8758}} \\ {{\rm 0.2706\; \; \; 0.2968\; \; \; 0.3066\; \; \; 0.5945\; \; \; 0.3039\; \; \; 0.9097\; \; \; 0.1923\; \; \; 0.7678}} \\ {{\rm 0.8174\; \; \; 0.3458\; \; \; 0.8938\; \; \; 0.6455\; \; \; 0.1174\; \; \; 0.1441\; \; \; 0.4414\; \; \; 0.5699}} \\ {{\rm 0.0249\; \; \; 0.0908\; \; \; 0.2651\; \; \; 0.6559\; \; \; 0.3363\; \; \; 0.2911\; \; \; 0.8178\; \; \; 0.3700}} \\ {{\rm 0.4353\; \; \; 0.2784\; \; \; 0.2792\; \; \; 0.9595\; \; \; 0.0716\; \; \; 0.2301\; \; \; 0.7262\; \; \; 0.2997}} \\ {{\rm 0.8223\; \; \; 0.1216\; \; \; 0.1137\; \; \; 0.2890\; \; \; 0.0377\; \; \; 0.0720\; \; \; 0.0247\; \; \; 0.0424}} \end{array}\right]\] 
\[{\it b}^{T} =[{\rm 0.9701\; ,\; 0.8656\; ,\; 0.836\; ,\; 0.7911\; ,\; 0.6082\; ,\; 0.4634\; ,\; 0.4634\; ,\; 0.3614}]\] 
\vskip 0.2in
\textbf{Test Problem 4}
\[f({\it x})=x_{1} -x_{2} -x_{3} -x_{1} x_{3} x_{5} +x_{1} x_{4} x_{6} +x_{2} x_{3} x_{7} -x_{2} x_{4} x_{8} \] 
\[{\it A}=\left[\begin{array}{l} {{\rm 0.4396\; \; \; 0.2216\; \; \; 0.1336\; \; \; 0.4682\; \; \; 0.2772\; \; \; 0.8983\; \; \; 0.0122\; \; \; 0.9332}} \\ {{\rm 0.9873\; \; \; 0.4027\; \; \; 0.3235\; \; \; 0.7378\; \; \; 0.1593\; \; \; 0.2108\; \; \; 0.9007\; \; \; 0.5111}} \\ {{\rm 0.8249\; \; \; 0.8397\; \; \; 0.0379\; \; \; 0.5054\; \; \; 0.5326\; \; \; 0.0086\; \; \; 0.5167\; \; \; 0.5803}} \\ {{\rm 0.6048\; \; \; 0.3073\; \; \; 0.2749\; \; \; 0.8800\; \; \; 0.9793\; \; \; 0.8859\; \; \; 0.0555\; \; \; 0.4956}} \\ {{\rm 0.3716\; \; \; 0.9327\; \; \; 0.4145\; \; \; 0.9292\; \; \; 0.9334\; \; \; 0.8875\; \; \; 0.1859\; \; \; 0.4239}} \\ {{\rm 0.2154\; \; \; 0.6681\; \; \; 0.8953\; \; \; 0.9269\; \; \; 0.9579\; \; \; 0.6790\; \; \; 0.0914\; \; \; 0.4114}} \\ {{\rm 0.1082\; \; \; 0.2890\; \; \; 0.0888\; \; \; 0.0474\; \; \; 0.1065\; \; \; 0.1732\; \; \; 0.0550\; \; \; 0.0853}} \\ {{\rm 0.0141\; \; \; 0.0057\; \; \; 0.0098\; \; \; 0.0094\; \; \; 0.0011\; \; \; 0.8573\; \; \; 0.0025\; \; \; 0.0104}} \end{array}\right]\] 
\[{\it b}^{T} =[{\rm 0.8964\; ,\; 0.788\; ,\; 0.788\; ,\; 0.6414\; ,\; 0.6414\; ,\; 0.6414\; ,\; 0.1439\; ,\; 0.0152}]\] 
\vskip 0.2in
\textbf{Test Problem 5}
\[f({\it x})=x_{1} x_{2} x_{3} x_{4} x_{5} -x_{6} x_{7} x_{8} +x_{9} x_{10} \] 
\[A=\left[\begin{array}{l} {{\rm 0.0600\; \; \; 0.2511\; \; \; 0.9209\; \; \; 0.2276\; \; \; 0.9494\; \; \; 0.2828\; \; \; 0.3557\; \; \; 0.7488\; \; \; 0.2268\; \; \; 0.8434}} \\ {{\rm 0.7095\; \; \; 0.4247\; \; \; 0.7763\; \; \; 0.3525\; \; \; 0.1900\; \; \; 0.8615\; \; \; 0.8905\; \; \; 0.5769\; \; \; 0.6407\; \; \; 0.9594}} \\ {{\rm 0.7882\; \; \; 0.9146\; \; \; 0.9574\; \; \; 0.8091\; \; \; 0.7355\; \; \; 0.7893\; \; \; 0.2965\; \; \; 0.2424\; \; \; 0.4028\; \; \; 0.3092}} \\ {{\rm 0.7316\; \; \; 0.9271\; \; \; 0.1748\; \; \; 0.5008\; \; \; 0.2375\; \; \; 0.0025\; \; \; 0.2371\; \; \; 0.9715\; \; \; 0.6156\; \; \; 0.7659}} \\ {{\rm 0.6527\; \; \; 0.6507\; \; \; 0.0047\; \; \; 0.2099\; \; \; 0.2427\; \; \; 0.7713\; \; \; 0.2320\; \; \; 0.6045\; \; \; 0.9449\; \; \; 0.2445}} \\ {{\rm 0.4446\; \; \; 0.7154\; \; \; 0.9425\; \; \; 0.5271\; \; \; 0.3922\; \; \; 0.5485\; \; \; 0.1432\; \; \; 0.3793\; \; \; 0.3806\; \; \; 0.0079}} \\ {{\rm 0.4234\; \; \; 0.7731\; \; \; 0.7489\; \; \; 0.2736\; \; \; 0.3397\; \; \; 0.6579\; \; \; 0.2917\; \; \; 0.7944\; \; \; 0.0262\; \; \; 0.3958}} \\ {{\rm 0.1634\; \; \; 0.6629\; \; \; 0.6402\; \; \; 0.3769\; \; \; 0.4384\; \; \; 0.6023\; \; \; 0.2094\; \; \; 0.0423\; \; \; 0.5819\; \; \; 0.5587}} \end{array}\right]\] 
\[{\it b}^{T} =[{\rm 0.9\; ,\; 0.7898\; ,\; 0.7898\; ,\; 0.6392\; ,\; 0.6392\; ,\; 0.5864\; ,\; 0.5864\; ,\; 0.5864}]\] 
\vskip 0.2in
\textbf{Test Problem 6}
\[f({\it x})=x_{1} +2x_{2} +4x_{5} +e^{x_{1} x_{4} x_{6} } -x_{7} x_{8} e^{2x_{9} -x_{10} } \] 
\[{\it A}=\left[\begin{array}{l} {{\rm 0.3929\; \; \; 0.9766\; \; \; 0.6832\; \; \; 0.5287\; \; \; 0.3723\; \; \; \; 0.8749\; \; \; 0.0264\; \; \; 0.9018\; \; \; 0.6531\; \; \; 0.9819}} \\ {{\rm 0.3702\; \; \; 0.1480\; \; \; 0.8312\; \; \; 0.4065\; \; \; 0.1637\; \; \; \; 0.4962\; \; \; 0.9869\; \; \; 0.0881\; \; \; 0.621\; \; \; \; \; 0.3531}} \\ {{\rm 0.2097\; \; \; 0.7964\; \; \; 0.3923\; \; \; 0.4738\; \; \; 0.2785\; \; \; \; 0.1016\; \; \; 0.2955\; \; \; 0.3064\; \; \; 0.9609\; \; \; 0.7242}} \\ {{\rm 0.3012\; \; \; 0.6326\; \; \; 0.1887\; \; \; 0.9143\; \; \; 0.1486\; \; \; \; 0.4006\; \; \; 0.4166\; \; \; 0.8941\; \; \; 0.5663\; \; \; 0.0352}} \\ {{\rm 0.6560\; \; \; 0.2583\; \; \; 0.1\; \; \; \; \; \; \; \; \; 0.8502\; \; \; 0.9685\; \; \; \; 0.8324\; \; \; 0.959\; \; \; \; \; 0.4153\; \; \; 0.5783\; \; \; 0.347}} \\ {{\rm 0.6323\; \; \; 0.3277\; \; \; 0.3326\; \; \; 0.2621\; \; \; 0.9914\; \; \; \; 0.6775\; \; \; 0.3566\; \; \; 0.6403\; \; \; 0.3587\; \; \; 0.1329}} \\ {{\rm 0.0060\; \; \; 0.9824\; \; \; 0.0962\; \; \; 0.1946\; \; \; 0.7119\; \; \; \; 0.4264\; \; \; 0.0015\; \; \; 0.2420\; \; \; 0.1303\; \; \; 0.0409}} \\ {{\rm 0.0841\; \; \; 0.8233\; \; \; 0.0659\; \; \; 0.1416\; \; \; 0.1047\; \; \; \; 0.3487\; \; \; 0.1516\; \; \; 0.0203\; \; \; 0.0451\; \; \; 0.0250}} \\ {{\rm 0.0462\; \; \; 0.4195\; \; \; 0.0872\; \; \; 0.0782\; \; \; 0.3259\; \; \; \; 0.4444\; \; \; 0.0940\; \; \; 0.3063\; \; \; 0.0446\; \; \; 0.0207}} \end{array}\right]\] 
\[{\it b}^{T} =[{\rm 0.9264\; ,\; 0.7977\; ,\; 0.7389\; ,\; 0.5941\; ,\; 0.5941\; ,\; 0.4387\; ,\; 0.2327\; ,\; 0.2327\; ,\; 0.2327}]\] 
\vskip 0.2in
\textbf{Test Problem 7}
\[f({\it x})=\sum _{k=1}^{9}[100(x_{k+1} -x_{k}^{2} )^{2} +(1-x_{k} )^{2} ] \] 
\[{\it A}=\left[\begin{array}{l} {{\rm 0.812\; \; \; \; \; 0.6281\; \; \; 0.063\; \; \; \; \; 0.9803\; \; \; 0.5762\; \; \; 0.0276\; \; \; 0.7127\; \; \; 0.7648\; \; \; 0.8193\; \; \; 0.6619}} \\ {{\rm 0.2792\; \; \; 0.8521\; \; \; 0.3791\; \; \; 0.2889\; \; \; 0.7784\; \; \; 0.9202\; \; \; 0.2201\; \; \; 0.3630\; \; \; 0.7759\; \; \; 0.5705}} \\ {{\rm 0.5447\; \; \; 0.4706\; \; \; 0.8592\; \; \; 0.0901\; \; \; 0.1094\; \; \; 0.3421\; \; \; 0.2168\; \; \; 0.6405\; \; \; 0.9930\; \; \; 0.8925}} \\ {{\rm 0.1321\; \; \; 0.817\; \; \; \; \; 0.0413\; \; \; 0.4569\; \; \; 0.7193\; \; \; 0.2761\; \; \; 0.8406\; \; \; 0.1417\; \; \; 0.7484\; \; \; 0.2615}} \\ {{\rm 0.7253\; \; \; 0.5051\; \; \; 0.8847\; \; \; 0.2749\; \; \; 0.4410\; \; \; 0.4622\; \; \; 0.5527\; \; \; 0.1059\; \; \; 0.1547\; \; \; 0.8351}} \\ {{\rm 0.4919\; \; \; 0.2471\; \; \; 0.9033\; \; \; 0.0069\; \; \; 0.1081\; \; \; 0.1483\; \; \; 0.0753\; \; \; 0.0198\; \; \; 0.4540\; \; \; 0.3356}} \\ {{\rm 0.3794\; \; \; 0.5373\; \; \; 0.0599\; \; \; 0.0794\; \; \; 0.1231\; \; \; 0.1716\; \; \; 0.1662\; \; \; 0.0796\; \; \; 0.1595\; \; \; 0.8758}} \end{array}\right]\] 
\[{\it b}^{T} =[{\rm 0.9303\; ,\; 0.7619\; ,\; 0.6297\; ,\; 0.5097\; ,\; 0.4705\; ,\; 0.2733\; ,\; 0.2619}]\] 
\vskip 0.2in
\textbf{Test Problem 8}
\[f({\it x})=-\, 0.5\, (x_{1} x_{4} -x_{2} x_{3} +x_{2} x_{6} -x_{5} x_{6} +x_{4} x_{5} -x_{6} x_{7} +x_{8} x_{10} -x_{9} x_{10} )\] 
\[{\it A}=\left[\begin{array}{l} {{\rm 0.2538\; \; \; 0.9943\; \; \; 0.5048\; \; \; 0.5869\; \; \; 0.1514\; \; \; 0.5405\; \; \; 0.1415\; \; \; 0.5711\; \; \; 0.4177\; \; \; 0.1420}} \\ {{\rm 0.2186\; \; \; 0.6899\; \; \; 0.1047\; \; \; 0.8187\; \; \; 0.7933\; \; \; 0.9628\; \; \; 0.7216\; \; \; 0.0617\; \; \; 0.1854\; \; \; 0.1476}} \\ {{\rm 0.4523\; \; \; 0.6951\; \; \; 0.8131\; \; \; 0.8095\; \; \; 0.1097\; \; \; 0.0215\; \; \; 0.7911\; \; \; 0.3582\; \; \; 0.635\; \; \; \; \; 0.6447}} \\ {{\rm 0.6647\; \; \; 0.4005\; \; \; 0.0798\; \; \; 0.8146\; \; \; 0.0883\; \; \; 0.1039\; \; \; 0.9671\; \; \; 0.4607\; \; \; 0.0305\; \; \; 0.9200}} \\ {{\rm 0.8746\; \; \; 0.1981\; \; \; 0.4623\; \; \; 0.7306\; \; \; 0.9437\; \; \; 0.3698\; \; \; 0.1068\; \; \; 0.5779\; \; \; 0.2977\; \; \; 0.6884}} \\ {{\rm 0.0370\; \; \; 0.1628\; \; \; 0.0232\; \; \; 0.0991\; \; \; 0.4626\; \; \; 0.0690\; \; \; 0.0706\; \; \; 0.0941\; \; \; 0.7398\; \; \; 0.0947}} \\ {{\rm 0.7203\; \; \; 0.0123\; \; \; 0.0470\; \; \; 0.0132\; \; \; 0.0576\; \; \; 0.0886\; \; \; 0.7022\; \; \; 0.5164\; \; \; 0.9357\; \; \; 0.0713}} \end{array}\right]\] 
\[{\it b}^{T} =[{\rm 0.9718\; ,\; 0.878\; ,\; 0.7243\; ,\; 0.568\; ,\; 0.568\; ,\; 0.1984\; ,\; 0.1006}]\] 
\vskip 0.2in
\textbf{Test Problem 9}
\[f({\it x})=e^{x_{1} x_{2} +x_{3} x_{6} +x_{7} x_{9} } -0.5\, (x_{1}^{3} +x_{2}^{3} +x_{8}^{3} +x_{10}^{3} +1)^{2} \] 
\[{\it A}=\left[\begin{array}{l} {{\rm 0.8011\; \; \; 0.4018\; \; \; 0.7035\; \; \; 0.9869\; \; \; 0.1705\; \; \; 0.9656\; \; \; 0.9832\; \; \; 0.8364\; \; \; 0.7133\; \; \; 0.7508}} \\ {{\rm 0.2569\; \; \; 0.4875\; \; \; 0.222\; \; \; \; \; 0.0709\; \; \; 0.9384\; \; \; 0.3757\; \; \; 0.9174\; \; \; 0.31\; \; \; \; \; \; \; 0.9603\; \; \; 0.408}} \\ {{\rm 0.8807\; \; \; 0.7017\; \; \; 0.7196\; \; \; 0.1238\; \; \; 0.0061\; \; \; 0.6642\; \; \; 0.9998\; \; \; 0.7438\; \; \; 0.1750\; \; \; 0.9458}} \\ {{\rm 0.7750\; \; \; 0.7051\; \; \; 0.1353\; \; \; 0.4416\; \; \; 0.2704\; \; \; 0.0217\; \; \; 0.0532\; \; \; 0.6864\; \; \; 0.6955\; \; \; 0.0207}} \\ {{\rm 0.7518\; \; \; 0.0550\; \; \; 0.9604\; \; \; 0.3945\; \; \; 0.1807\; \; \; 0.7580\; \; \; 0.1978\; \; \; 0.0644\; \; \; 0.2631\; \; \; 0.7684}} \\ {{\rm 0.9177\; \; \; 0.5041\; \; \; 0.9828\; \; \; 0.2669\; \; \; 0.2915\; \; \; 0.1256\; \; \; 0.0094\; \; \; 0.9482\; \; \; 0.034\; \; \; \; \; 0.6577}} \\ {{\rm 0.8391\; \; \; 0.2024\; \; \; 0.6265\; \; \; 0.2043\; \; \; 0.3186\; \; \; 0.0997\; \; \; 0.0483\; \; \; 0.0232\; \; \; 0.3402\; \; \; 0.1631}} \\ {{\rm 0.543\; \; \; \; \; 0.9321\; \; \; 0.5742\; \; \; 0.0956\; \; \; 0.3085\; \; \; 0.3283\; \; \; 0.0358\; \; \; 0.6781\; \; \; 0.1433\; \; \; 0.1817}} \\ {{\rm 0.2430\; \; \; 0.6933\; \; \; 0.9497\; \; \; 0.0620\; \; \; 0.0937\; \; \; 0.0165\; \; \; 0.1313\; \; \; 0.1493\; \; \; 0.1676\; \; \; 0.1608}} \\ {{\rm 0.2594\; \; \; 0.8109\; \; \; 0.9014\; \; \; 0.0829\; \; \; 0.0854\; \; \; 0.2434\; \; \; 0.0184\; \; \; 0.0365\; \; \; 0.1109\; \; \; 0.2225}} \end{array}\right]\] 
\[{\it b}^{T} =[{\rm 0.9758\; ,\; 0.9288\; ,\; 0.9288\; ,\; 0.6185\; ,\; 0.5077\; ,\; 0.4076\; ,\; 0.3434\; ,\; 0.3434\; ,\; 0.2977\; ,\; 0.2977}]\] 
\vskip 0.2in
\textbf{Test Problem 10}
\[f({\it x})=(x_{1} -1)^{2} +(x_{7} -1)^{2} +10\sum _{k=1}^{11}(10-k)(x_{k}^{2} -x_{k+1} )^{2}  \] 
\[{\it A}=\left[\begin{array}{l} {{\rm 0.0702\; \; \; 0.1745\; \; \; 0.5925\; \; \; 0.2656\; \; \; 0.8812\; \; \; 0.6833\; \; \; 0.003\; \; \; \; \; 0.5535\; \; \; 0.7155\; \; \; 0.8598\; \; \; 0.4763\; \; \; 0.9709}} \\ {{\rm 0.963\; \; \; \; \; 0.5863\; \; \; 0.1001\; \; \; 0.828\; \; \; \; \; 0.2939\; \; \; 0.5813\; \; \; 0.2774\; \; \; 0.5876\; \; \; 0.5299\; \; \; 0.9017\; \; \; 0.6051\; \; \; 0.8993}} \\ {{\rm 0.7901\; \; \; 0.5698\; \; \; 0.8892\; \; \; 0.5924\; \; \; 0.4591\; \; \; 0.1807\; \; \; 0.3446\; \; \; 0.5325\; \; \; 0.9014\; \; \; 0.4529\; \; \; 0.7917\; \; \; 0.345}} \\ {{\rm 0.6321\; \; \; 0.7162\; \; \; 0.6132\; \; \; 0.2433\; \; \; 0.6247\; \; \; 0.0607\; \; \; 0.9719\; \; \; 0.1374\; \; \; 0.3581\; \; \; 0.1596\; \; \; 0.8537\; \; \; 0.3707}} \\ {{\rm 0.8638\; \; \; 0.8306\; \; \; 0.6024\; \; \; 0.9979\; \; \; 0.2076\; \; \; 0.5049\; \; \; 0.7025\; \; \; 0.1535\; \; \; 0.9984\; \; \; 0.5087\; \; \; 0.1401\; \; \; 0.7828}} \\ {{\rm 0.2246\; \; \; 0.3024\; \; \; 0.5784\; \; \; 0.1239\; \; \; 0.5353\; \; \; 0.5777\; \; \; 0.2191\; \; \; 0.466\; \; \; \; \; 0.2639\; \; \; 0.1992\; \; \; 0.8516\; \; \; 0.0107}} \\ {{\rm 0.6065\; \; \; 0.5062\; \; \; 0.5623\; \; \; 0.1472\; \; \; 0.0783\; \; \; 0.9427\; \; \; 0.2715\; \; \; 0.4176\; \; \; 0.2902\; \; \; 0.1834\; \; \; 0.9231\; \; \; 0.0426}} \\ {{\rm 0.9605\; \; \; 0.0447\; \; \; 0.794\; \; \; \; \; 0.0357\; \; \; 0.1938\; \; \; 0.3113\; \; \; 0.2229\; \; \; 0.5738\; \; \; 0.0477\; \; \; 0.1005\; \; \; 0.9174\; \; \; 0.1864}} \\ {{\rm 0.0254\; \; \; 0.4711\; \; \; 0.3339\; \; \; 0.2237\; \; \; 0.0934\; \; \; 0.7857\; \; \; 0.2072\; \; \; 0.1740\; \; \; 0.1953\; \; \; 0.0672\; \; \; 0.2243\; \; \; 0.2333}} \\ {{\rm 0.0444\; \; \; 0.1702\; \; \; 0.5993\; \; \; 0.0805\; \; \; 0.1675\; \; \; 0.8279\; \; \; 0.0647\; \; \; 0.1368\; \; \; 0.1043\; \; \; 0.1891\; \; \; 0.1084\; \; \; 0.1822}} \end{array}\right]\] 
\[{\it b}^{T} =[{\rm 0.9615\; ,\; 0.9002\; ,\; 0.9002\; ,\; 0.9002\; ,\; 0.9002\; ,\; 0.4477\; ,\; 0.3132\; ,\; 0.3132\; ,\; 0.2893\; ,\; 0.2256}]\] 
\textbf{}
\vskip 0.2in
\noindent \textbf{Reference}
\vskip 0.2in
\noindent [1]. M.S. Bazaraa, H.D. Sherali, C.M. Shetty, Nonlinear programming: Theory and Algorithms, John Wiley and Sons, New York, NY, 2006.

\noindent [2]. M.M. Bourke, D.G. Fisher, Solution algorithms for fuzzy relational equations with max-product composition, Fuzzy Sets and Systems 94 (1998) 61-69.

\noindent [3]. C. W. Chang, B. S. Shieh, Linear optimization problem constrained by fuzzy max--min relation equations, Information Sciences 234 (2013) 71--79.

\noindent 

\noindent [4]. L. Chen, P. P. Wang, Fuzzy relation equations (ii): the branch-point-solutions and the categorized minimal solutions, Soft Computing 11 (2) (2007) 33-40.

\noindent [5]. S. Dempe, A. Ruziyeva, On the calculation of a membership function for the solution of a fuzzy linear optimization problem, Fuzzy Sets and Systems 188 (2012) 58-67.

\noindent [6]. F. Di Martino, V. Loia, S. Sessa, Digital watermarking in coding/decoding processes with fuzzy relation equations, Soft Computing 10(2006) 238-243.

\noindent [7]. A. Di Nola, S. Sessa, W. Pedrycz, E. Sanchez, Fuzzy relational equations and their applications in knowledge engineering, Dordrecht: Kluwer academic press, 1989.

\noindent [8]. M. Dorigo, V. Maniezzo, A. Colorni, Ant System: Optimization by a colony of cooperating agents, IEEE Transactions on Systems, Man and Cybernetics 26(1) (1996) 29-41.    

\noindent [9]. D. Dubey, S. Chandra, A. Mehra, Fuzzy linear programming under interval uncertainty based on IFS representation, Fuzzy Sets and Systems 188 (2012) 68-87.

\noindent [10]. A. ElSaid, F. El Jamiy, J. Higgins, B. Wild, T. Desell, Optimizing long short-term memory recurrent neural networks using ant colony optimization to predict turbine engine vibration, Applied Soft Computing 73 (2018) 969-991.

\noindent [11]. Y. R. Fan, G. H. Huang, A. L. Yang, Generalized fuzzy linear programming for decision making under uncertainty: Feasibility of fuzzy solutions and solving approach, Information Sciences 241 (2013) 12-27. 

\noindent [12]. S. C. Fang, G. Li, Solving fuzzy relation equations with a linear objective function, Fuzzy Sets and Systems 103(1999) 107-113. 

\noindent [13]. S. Freson,~B. De Baets, H. De Meyer, Linear optimization with bipolar max--min constraints, Information Sciences 234 (2013) 3--15.

\noindent [14]. A. Ghodousian, Optimization of linear problems subjected to the intersection of two fuzzy relational inequalities defined by Dubois-Prade family of t-norms, Information Sciences 503 (2019) 291--306.

\noindent [15].  A. Ghodousian, A. Babalhavaeji, An effcient genetic algorithm for solving nonlinear optimization problems defined with fuzzy relational equations and max-Lukasiewicz composition, Applied Soft Computing 69 (2018) 475--492.

\noindent [16]. A. Ghodousian, E. Khorram, Fuzzy linear optimization in the presence of the fuzzy relation inequality constraints with max-min composition, Information Sciences 178 (2008) 501-519.

\noindent [17]. A. Ghodousian, E. Khorram, Linear optimization with an arbitrary fuzzy relational inequality, Fuzzy Sets and Systems 206 (2012) 89-102.

\noindent [18]. A. Ghodousian, M. Naeeimib, A. Babalhavaeji, Nonlinear optimization problem subjected to fuzzy relational equations defined by Dubois-Prade family of t-norms, Computers \& Industrial Engineering 119 (2018) 167--180.

\noindent [19]. A. Ghodousian, M. Raeisian Parvari, A modified PSO algorithm for linear optimization problem subject to the generalized fuzzy relational inequalities with fuzzy constraints (FRI-FC), Information Sciences 418--419 (2017) 317--345.

\noindent [20]. F. F. Guo, Z. Q. Xia, An algorithm for solving optimization problems with one linear objective function and finitely many constraints of fuzzy relation inequalities, Fuzzy Optimization and Decision Making 5(2006) 33-47.

\noindent [21]. S. M. Guu, Y. K. Wu, Minimizing a linear objective function with fuzzy relation equation constraints, Fuzzy Optimization and Decision Making 12 (2002) 1568-4539.

\noindent [22]. S. M. Guu, Y. K. Wu, Minimizing an linear objective function under a max-t-norm fuzzy relational equation constraint, Fuzzy Sets and Systems 161 (2010) 285-297.

\noindent [23]. M. Higashi, G.J. Klir, Resolution of finite fuzzy relation equations, Fuzzy Sets and Systems 13 (1984) 65-82.

\noindent [24]. K. Krynicki, J. Jaen, E. Navarro, An ACO-based personalized learning technique in support of people with acquired brain injury, Applied Soft Computing 47 (2016) 316-331.

\noindent [25]. H. C. Lee, S. M. Guu, On the optimal three-tier multimedia streaming services, Fuzzy Optimization and Decision Making 2(1) (2002) 31-39.

\noindent [26]. P. K. Li, S. C. Fang, On the resolution and optimization of a system of fuzzy relational equations with sup-t composition, Fuzzy Optimization and Decision Making 7 (2008) 169-214.

\noindent [27]. P. Li, Y. Liu, Linear optimization with bipolar fuzzy relational equation constraints using lukasiewicz triangular norm, Soft Computing 18 (2014) 1399-1404.

\noindent [28]. P. Li, Y. Wang, A matrix approach to latticized linear programming with fuzzy-relatoion inequality constraints, IEEE Transactions on Fuzzy Systems 21 (2013) 781-788.

\noindent [29]. C.-H. Lin, A rough penalty genetic algorithm for constrained optimization, Inf. Sci. (NY) 241 (2013) 119--137.

\noindent [30]. J. L. Lin, Y. K. Wu, S. M. Guu, On fuzzy relational equations and the covering problem, Information Sciences 181 (2011) 2951-2963.

\noindent [31]. J. Liu, J. Liu, Applying multi-objective ant colony optimization algorithm for solving the unequal area facility layout problems, Applied Soft Computing 74 (2019) 167-189.

\noindent [32]. C. C. Liu, Y. Y. Lur, Y. K. Wu, Linear optimization of bipolar fuzzy relational equations with max-{\L}ukasiewicz composition, Information Sciences 360 (2016) 149--162.

\noindent [33]. J. Loetamonphong, S. C. Fang, Optimization of fuzzy relation equations with max-product composition, Fuzzy Sets and Systems 118 (2001) 509-517.

\noindent [34]. J. Lu, S. C. Fang, Solving nonlinear optimization problems with fuzzy relation equations constraints, Fuzzy Sets and Systems 119(2001) 1-20.

\noindent [35]. L. C. Lu, T. W. Yue, Mission-oriented ant-team ACO for min--max MTSP, Applied Soft Computing 76 (2019) 436-444.

\noindent [36]. J. Lv, X. Wang, M. Huang, ACO-inspired ICN Routing mechanism with Mobility support, Applied Soft Computing, 58 (2017) 427-440.

\noindent [37]. A. V. Markovskii, On the relation between equations with max-product composition and the covering problem, Fuzzy Sets and Systems 153 (2005) 261-273.

\noindent [38]. Z. Matusiewicz, J. Drewniak, Increasing continuous operations in fuzzy max-* equations and inequalities, Fuzzy Sets and Systems 232 (2013) 120-133.

\noindent [39]. M. Mizumoto, H. J. Zimmermann, Comparison of fuzzy reasoning method, Fuzzy Sets and Systems 8 (1982) 253-283.

\noindent [40]. H. Nobuhara, K. Hirota, F. Di Martino, W. Pedrycz, S. Sessa, Fuzzy relation equations for compression/decompression processes of color images in the RGB and YUV color spaces, Fuzzy Optimization and Decision Making 4 (2005) 235-246.

\noindent [41]. W. Pedrycz, Granular Computing: Analysis and Design of Intelligent Systems, CRC Press, Boca Raton, 2013.

\noindent [42]. W. Pedrycz, On generalized fuzzy relational equations and their applications, Journal of Mathematical Analysis and Applications 107 (1985) 520-536.

\noindent [43]. W. Pedrycz, A. V. Vasilakos, Modularization of fuzzy relational equations, Soft Computing 6(2002) 3-37.

\noindent [44]. K. Peeva, Resolution of fuzzy relational equations-methods, algorithm and software with applications, Information Sciences 234 (2013) 44-63.

\noindent [45]. S. Perez-Carabaza, E. Besada-Portas, J. A. Lopez-Orozco, J. M. de la Cruz, Ant colony optimization for multi-UAV minimum time search in uncertain domains, Applied Soft Computing 62 (2018) 789-806.

\noindent [46]. I. Perfilieva, Fuzzy function as an approximate solution to a system of fuzzy relation equations, Fuzzy Sets and Systems 147(2004) 363-383.

\noindent [47]. I. Perfilieva, Finitary solvability conditions for systems of fuzzy relation equations, Information Sciences 234 (2013)29-43.

\noindent [48]. I. Perfilieva, V. Novak, System of fuzzy relation equations model of IF-THEN rules, Information Sciences 177 (16) (2007) 3218-3227.

\noindent [49]. D. Praveen Kumar, A. Tarach, A. Chandra Sekhara Rao, ACO-based mobile sink path determination for wireless sensor networks under non-uniform data constraints, Applied Soft Computing 69 (2018) 528-540.

\noindent [50]. X. B. Qu, X. P. Wang, Man-hua. H. Lei, Conditions under which the solution sets of fuzzy relational equations over complete Brouwerian lattices form lattices, Fuzzy Sets and Systems 234 (2014) 34-45.

\noindent [51]. W. Rudin,~Principles of Mathematical Analysis, New York: McGraw-Hill, 1976.

\noindent [52]. E. Sanchez, Resolution of composite fuzzy relation equations, Inf. Control 30(1976) 38-48.

\noindent [53]. E. Sanchez, Solution in composite fuzzy relation equations: application to medical diagnosis in Brouwerian logic, in: M.M. Gupta. G.N. Saridis, B.R. Games (Eds.), Fuzzy Automata and Decision Processes, North-Holland, New York, 1977, pp. 221-234.

\noindent [54]. B. S. Shieh, Minimizing a linear objective function under a fuzzy max-t-norm relation equation constraint, Information Sciences 181 (2011) 832-841.

\noindent [55]. K. Socha, M. Dorigo, Ant colony optimization for continuous domain, European Journal of operational Research 185 (2008) 1155-1173.

\noindent [56]. F. Sun, X. P. Wang, x. B. Qu, Minimal join decompositions and their applications to fuzzy relation equations over complete Brouwerian lattices, Information Sciences 224 (2013) 143-151. 

\noindent [57]. P. Z. Wang, Latticized linear programming and fuzzy relation inequalities, Journal of Mathematical Analysis and Applications 159(1) (1991) 72-87.  

\noindent [58]. Y. K. Wu, Optimization of fuzzy relational equations with max-av composition, Information Sciences 177 (2007) 4216-4229.

\noindent [59]. Y. K. Wu, S. M. Guu, A note on fuzzy relation programming problems with max-strict-t-norm composition,  Fuzzy Optimization and Decision Making 3(2004) 271-278.

\noindent [60]. Y. K. Wu, S. M. Guu, An efficient procedure for solving a fuzzy relation equation with max-Archimedean t-norm composition, IEEE Transactions on Fuzzy Systems 16 (2008) 73-84.

\noindent [61]. Y. K. Wu, S. M. Guu, Minimizing a linear function under a fuzzy max-min relational equation constraints, Fuzzy Sets and Systems 150 (2005) 147-162.

\noindent [62]. Y. K. Wu, S. M. Guu, J. Y. Liu, Reducing the search space of a linear fractional programming problem under fuzzy relational equations with max-Archimedean t-norm composition, Fuzzy Sets and Systems 159 (2008) 3347-3359.

\noindent [63]. Q. Q. Xiong, X. P. Wang, Fuzzy relational equations on complete Brouwerian lattices, Information Sciences 193 (2012) 141-152.

\noindent [64]. Y. Yan, H. S. Sohn, G. Reyes, A modified ant system to achieve better balance between intensification and diversification for the traveling salesman problem, Applied Soft Computing 60 (2017) 256-267.

\noindent [65]. X. P. Yang, X. G. Zhou, B. Y. Cao, Latticized linear programming subject to max-product fuzzy relation inequalities with application in wireless communication, Information Sciences\textbf{ }358--359 (2016) 44--55.

\noindent [66]. X. P. Yang, X. G. Zhou, B. Y. Cao, Min-max programming problem subject to addition-min fuzzy relation inequalities, IEEE Transactions on Fuzzy Systems\textbf{ }24 (1) (2016) 111--119.

\noindent [67].  X. P. Yang, X. G. Zhou, B. Y. Cao, Single-variable term semi-latticized fuzzy relation geometric programming with max-product operator, Information Sciences 325 (2015) 271-287.

\noindent \textbf{}

\end{document}